\documentclass{IEEEtran}
\usepackage{cite}
\usepackage{amsmath,amssymb,amsfonts,amsthm}
\usepackage{algorithmic}
\usepackage{graphicx}
\usepackage{textcomp}

\usepackage{xurl} % allow arbitrary line breaks in a formatted URL string
\usepackage{booktabs}
\usepackage{multirow}
\usepackage{lipsum}

\usepackage[caption=false,font=normalsize,labelfont=sf,textfont=sf]{subfig}
\newtheorem{definition}{Definition}
\def\BibTeX{{\rm B\kern-.05em{\sc i\kern-.025em b}\kern-.08em
    T\kern-.1667em\lower.7ex\hbox{E}\kern-.125emX}}

\usepackage{bm}

\makeatother

\def\BibTeX{{\rm B\kern-.05em{\sc i\kern-.025em b}\kern-.08em
    T\kern-.1667em\lower.7ex\hbox{E}\kern-.125emX}}

\begin{document}

\title{On Electric Vehicle Energy Demand Forecasting and the Effect of Federated Learning}
\author{%
    Andreas Tritsarolis, Gil Sampaio, Nikos Pelekis, Yannis Theodoridis%
    \thanks{A. Tritsarolis (andrewt@unipi.gr) and Y. Theodoridis (ytheod@unipi.gr) are with the Department of Informatics, University of Piraeus, Piraeus, Greece}%
    \thanks{G. Sampaio (gil.s.sampaio@inesctec.pt) is with the Center for Power and Energy Systems, INESC TEC, Porto, Portugal}%
    \thanks{N. Pelekis (npelekis@unipi.gr) is with the Department of Statistics and Insurance Science, University of Piraeus, Piraeus, Greece}%
}

\maketitle

\begin{abstract}
    The wide spread of new energy resources, smart devices, and demand side management strategies has motivated several analytics operations, from infrastructure load modeling to user behavior profiling. Energy Demand Forecasting (EDF) of Electric Vehicle Supply Equipments (EVSEs) is one of the most critical operations for ensuring efficient energy management and sustainability, since it enables utility providers to anticipate energy/power demand, optimize resource allocation, and implement proactive measures to improve grid reliability. However, accurate EDF is a challenging problem due to external factors, such as the varying user routines, weather conditions, driving behaviors, unknown state of charge, etc. Furthermore, as concerns and restrictions about privacy and sustainability have grown, training data has become increasingly fragmented, resulting in distributed datasets scattered across different data silos and/or edge devices, calling for federated learning solutions. 
    In this paper, we investigate different well-established time series forecasting methodologies to address the EDF problem, from statistical methods (the ARIMA family) to traditional machine learning models (such as XGBoost) and deep neural networks (GRU and LSTM).
    We provide an overview of these methods through a performance comparison over four real-world EVSE datasets, evaluated under both centralized and federated learning paradigms, focusing on the trade-offs between forecasting fidelity, privacy preservation, and energy overheads. Our experimental results demonstrate, on the one hand, the superiority of gradient boosted trees (XGBoost) over statistical and NN-based models in both prediction accuracy and energy efficiency and, on the other hand, an insight that Federated Learning-enabled models balance these factors, offering a promising direction for decentralized energy demand forecasting.
\end{abstract}


\section{Introduction}\label{sec:introduction}
    The increasing adoption of Distributed Energy Resources (DERs), such as rooftop photovoltaics, smart home systems, and Electric Vehicles (EVs), is reshaping power grid operations. While these technologies promote cleaner and more efficient energy systems, their decentralized and variable nature introduces operational challenges. For instance, uncoordinated EV charging during peak demand periods can strain grid infrastructure, reduce efficiency (e.g., increased losses), and potentially increase greenhouse gas (GHG) emissions due to reliance on fossil-fuel generation \cite{en14082233,LI2023104708,8307186} when renewables have to be curtailed.

    Globally, policymakers aim to reduce emissions while maintaining supply security and economic competitiveness \cite{MATTHEW2017121}. As the EV population grows, risks such as severe voltage drops and transformer / power-line overloads become more likely. Nevertheless, the inherent variability of EVs, arising from irregular charging patterns and battery storage capabilities, motivates research for energy distribution strategies that reduce generation costs, and mitigate technical issues like peak-demand spikes and local grid congestion without compromising grid limits \cite{6881739}.
    
    Smart charging is central to managing the growing population of EVs. Its purpose lies in planning and scheduling charging sessions based on real-time feedback from both the grid and vehicles \cite{SADEGHIAN2022105241,Menegatti2023}. Without such coordination, unregulated charging can lead to distribution grid congestion and suboptimal cost outcomes. Smart charging, conversely, aligns consumption with grid conditions and price signals, allowing EVs to mitigate their own adverse impacts and even contribute positively to grid stability and renewable integration \cite{GONZALEZGARRIDO2019381,PECASLOPES201024}.

    Accurate Energy Demand Forecasting (EDF) is an emerging, yet critical research field towards ensuring grid stability and sustainability \cite{app9091723}. Unlike household or industrial loads, EV demand is highly erratic, influenced by geographical, climatic, and behavioral factors. Existing forecasting models often rely on oversimplified assumptions, such as perfect foresight of charging parameters (e.g., known initial state of charge) \cite{8307186} or price-responsive users \cite{LI2023104708}, which may not reflect real-world conditions. These limitations highlight the need for robust forecasting tools capable of handling uncertainty in EV load profiles and dynamic grid constraints.

    In this paper, we study EDF from the perspective of time series forecasting, and conduct an experimental comparison across several well-established methodologies ranging from baseline statistical methods (e.g., ARIMA) to traditional machine learning models (such as XGBoost) and deep neural networks (GRU and LSTM). Furthermore, to address the decentralized nature of the grid, particularly from EV Supply Equipment (EVSE), we extend our comparison to the Federated Learning (FL) paradigm \cite{DBLP:conf/aistats/McMahanMRHA17,DBLP:series/synthesis/2019YangLCKCY,DBLP:journals/ftml/KairouzMABBBBCC21}, and assess the trade-offs between accuracy, privacy, and energy consumption overhead, compared to the (standard) centralized approach. Through this analysis, our goal is to provide actionable insights for policymakers and grid operators on balancing sustainability, reliability, and economic efficiency in the era of rapid electrification. In summary, the main contributions of this work are as follows: 

    % Contribution(s)
    \begin{itemize}
        \item We examine well-established data-driven methods from statistics and Machine Learning (ML) in order to address the EDF problem 
        \item We provide comparative results on prediction accuracy for the next 12-hour interval.
        \item We discuss the accuracy vs. privacy trade-off of the centralized vs. federated learning paradigms, and assess the associated carbon-emission overhead.
    \end{itemize}
    In addition, we provide our study as open-source\footnote{The corresponding source code used in our experimental study is available at: \url{https://github.com/DataStories-UniPi/FedEDF}} in order for the researchers and practitioners in the field to take the most benefit from it. To our knowledge,  this work is the first to jointly analyze the accuracy vs. energy trade-offs of centralized and federated time series forecasters on real-world EVSE datasets while treating the problem through an intermittent-and-lumpy forecasting lens \cite{DBLP:conf/hicss/KieferGBD21}.
    
    % Sections Description    
    The rest of the paper is organized as follows. Section \ref{sec:RelatedWork} discusses related work. Section \ref{sec:Background_and_Definitions} formulates the problem at hand and presents the architecture of our methodological framework. Section \ref{sec:experiments} presents our experimental study and discusses the main findings. Finally, Section \ref{sec:conclusion} concludes the paper, also giving hints for future work.

\section{Related Work}\label{sec:RelatedWork}    
    % Include Rel. Work on FedEDF
    \subsection{Energy Demand Forecasting}\label{subsec:RelatedWork_EDF}    
        Considering the EDF problem, current state of the art includes an adequate number of research works \cite{en14082233}. 
        Zhu et al. \cite{app9091723} exploit on baseline recurrent neural network (RNN)-based algorithms and propose a Gated Recurrent Unit (GRU)-based approach for performing EDF on hourly resolution.
        Divina et al. \cite{Divina2019} present a comparative study of different time series forecasting techniques for predicting energy consumption in non-residential buildings up to 24 hours into the future.
        Yi et al. \cite{doi:10.1080/15472450.2021.1966627} perform an experimental comparison on ML models for EDF and propose a sequence-to-sequence (Seq2Seq) LSTM-based architecture for predicting EV Charging Stations (EVCSs) demand, in terms of occupancy, within either 1 (single-step prediction) or 5 (multi-step prediction) months.
        Unterluggauger et al. \cite{https://doi.org/10.1049/els2.12028} perform a similar case study, however their Seq2Seq LSTM-based approach focuses on energy demand forecasting at a finer (15-minute) resolution, up to 24 hours in the future.

        Towards more recent works, Skaloumpakas et al. \cite{Skaloumpakas2024} exploit on Gradient Boosted Trees for predicting energy demand of an EVCS within the next 24 hours, whereas 
        Mohammad et al. \cite{Mohammad2023} propose two encoder-decoder models based on convolutional long short-term memory networks (ConvLSTM) and bidirectional ConvLSTM (BiConvLSTM) in combination with the standard long short-term memory (LSTM) network, to predict the energy demand at a higher (24-hour) resolution, up to 168 hours (1 week).

    % Include Rel. Work on FL
    \subsection{Federated Learning}\label{subsec:RelatedWork_FedEDF}
        \begin{figure*}[!ht]
            \centering
            \includegraphics[width=\textwidth]{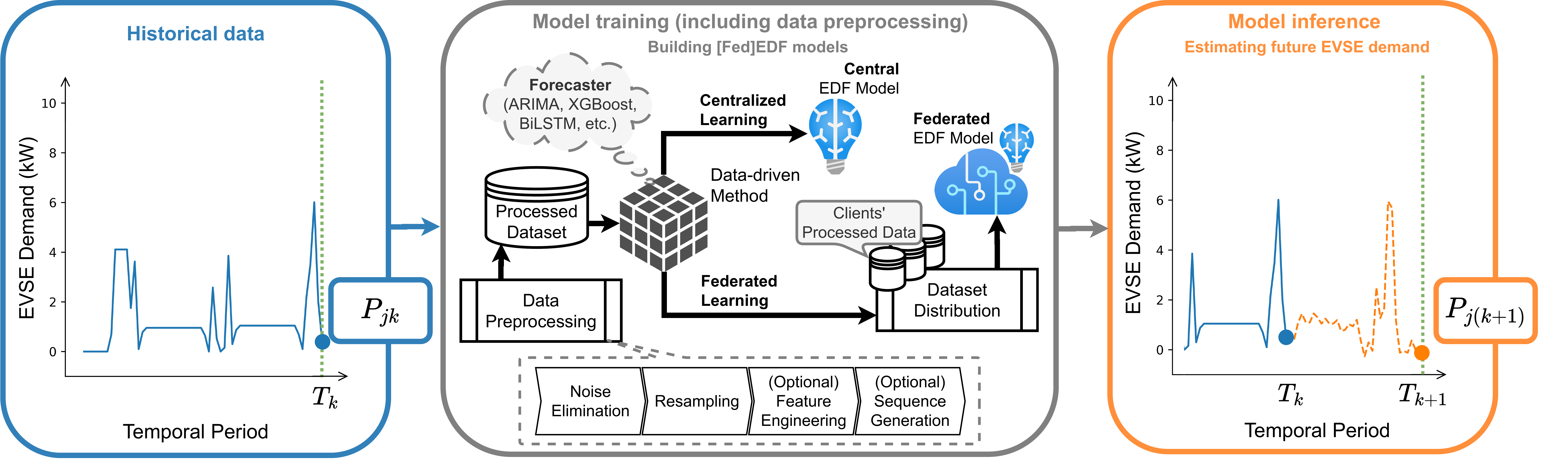}
            \caption{Overview of our [Fed]EDF forecasting framework.}
            \label{fig:EnergyDemandForecasting-testbed}
        \end{figure*}
        
        While distributed ML \cite{DBLP:conf/www/LiuCW17} can help scale up the training process across multiple computational nodes, it can only be used on centralized data. On the other hand, FL trains centralized models using decentralized data \cite{DBLP:conf/aistats/McMahanMRHA17,DBLP:series/synthesis/2019YangLCKCY,DBLP:journals/ftml/KairouzMABBBBCC21}, as such, (cross-device) FL algorithms are primarily geared towards data privacy. 
        
        In particular for federated EDF, Saputra et al. \cite{DBLP:conf/globecom/SaputraHNDMS19} propose FEDL, an FL approach for predicting energy demand on each vehicle that plugs into the EVCS. To improve forecasting accuracy, they propose an extension of FEDL in which the EVCS are clustered into similar groups (with respect to location) before applying FEDL.  Further following this line of research, through a baseline LSTM-based neural network, Taïk et al. \cite{DBLP:conf/icc/TaikC20} demonstrate the soundness of FL towards EDF in terms of bandwidth gain and privacy preservation, as well as the effect of personalisation of the global model on the accuracy of the edge devices' data. 
        
        In a similar line of research, Thorgeirsson et al. \cite{Thorgeirsson2021} apply an extension of the FedAvg \cite{DBLP:journals/corr/McMahanMRA16} algorithm to learn probabilistic neural networks and linear regression models. Their experimental study on two FL aggregation schemes, namely FedAvg and FedAvg-Gaussian, demonstrate the advantages of probabilistic EDF, as well as their effect on the driving behavior in terms of battery utilisation and effective driving range. 
        
        Towards more recent works, Menegatti et al. \cite{Menegatti2023}, propose a novel consensus-based protocol for peer-to-peer FL, and assess its performance compared to centralized and distributed approach over a baseline LSTM neural network. Their experimental study over an open energy demand dataset, demonstrates the soundness of their approach, by achieving comparable performance to the centralized setting.

        In contrast to the above related work, we perform a systematic, side-by-side evaluation of established forecasting models in both centralized and federated settings over four real-world EVSE datasets, reporting not only the prediction accuracy of the models but also the training and inference energy footprint of each scheme.

%% Background and Definitions
\section{Problem Formulation and Proposed Methodology}\label{sec:Background_and_Definitions}
    In this section, we formulate the Energy Demand Forecasting (EDF) problem and describe our proposed methodology.
            
    \subsection{Problem Definition}\label{subsec:ProblemDefinition}
        Before we proceed with the actual formulation of the problem, we provide some preliminary definitions.

        \begin{definition}\label{def:EnergyTransaction}
            (Energy Transaction). An energy transaction $et_{ij}$ between an EV $i$ and an EV supply equipment (EVSE) $j$ is defined as $e_{ij} = \{ t_{start}^{ij}, t_{end}^{ij}, e^{ij}, p^{ij} \}$, where $e^{ij}$ is the total energy consumed (in kWh) during the temporal interval $\lbrack t_{start}^{ij}, t_{end}^{ij} \rbrack$ and $p^{ij}$ is the corresponding average charging power (in kW), or,
            $$ p^{ij} = \dfrac{e^{ij}}{t_{end}^{ij} - t_{start}^{ij}}$$
        \end{definition}
        
        \begin{definition}\label{def:EnergyDemand}
            (Energy Demand). Given a temporal axis $T$ with fixed sampling rate $sr_{freq}$, the energy demand $P_{jk}$ of EVSE $j$ during period $T_k$, multiple of $sr_{freq}$, is defined as the sum of the average charging power $p^{ij}$ of each transaction $e_{ij}$ weighted by the overlap of their corresponding interval $\lbrack t_{start}^{ij}, t_{end}^{ij} \rbrack$ with respect to $T_k$, or, 
            $$P_{jk} = \sum_{i}{ p^{ij} \dfrac{\lvert T_k \bigcap [t_{start}^{ij}, t_{end}^{ij}] \rvert}{\lvert T_k \rvert}},$$
            where $\lvert \cdot \rvert$ and $\bigcap$ is the length and intersection operations, respectively, as defined in Allen's temporal algebra \cite{DBLP:journals/cacm/Allen83}.
        \end{definition}
    
        \begin{definition}\label{def:EnergyDemandForecasting}
            (Energy Demand Forecasting). Given a dataset $D$ of EVSEs' historic power demand in the form of time series over a temporal axis $T$ and their current demand sequence $P_{jk}$ up to period $T_k$, the goal is to train a data-driven model over $D$, which is able to estimate EVSEs' future energy demand in the next period $T_{k+1}$, or, $P_{j(k+1)}$.
        \end{definition}

        If we recall Figure \ref{fig:EnergyDemandForecasting-testbed}, it provides an illustration of the methodological framework developed to support Definition \ref{def:EnergyDemandForecasting}. 
        At first, we aggregate the raw transaction records of an EVSE (cf., Definition \ref{def:EnergyTransaction}), into a time series of average charging power per fixed-size temporal interval, thereby obtaining the energy demand sequence described in Definition \ref{def:EnergyDemand}. Afterwards, in the training step, we feed our Forecaster (e.g., ARIMA, XGBoost, LSTM, etc.) with a pre-processed historical dataset $D$ and train the respective EDF model, in two variations (centralized and federated, where possible). Finally, in the inference step, once we know the energy demand sequence $P_{jk}$ of EVSE $j$ up to period $T_{k}$, we estimate the anticipated energy demand of this EVSE in the next period $T_{k+1}$. The details of our methodological framework and the respective workflow are provided in the sections that follow.

    \subsection{Our Methodology for Training EDF Models}\label{subsec:Methodology}           
        The EDF task falls under the category of intermittent and lumpy time series forecasting methods \cite{DBLP:conf/hicss/KieferGBD21}, as EVSEs' energy demand occurs sporadically (i.e., intermittent) with significant variance in terms of energy output (i.e., lumpy). Such demand patterns are especially difficult to forecast \cite{DBLP:conf/fskd/XuWS12}, thereby we need to exercise caution in order to avoid model over-fitting on the non-demand periods. To address the EDF task, we experiment with three method families:

        \begin{itemize}
            \item Statistical Autoregressive Methods (ARIMA and variations)
            \item eXtreme Gradient Boosted Trees (XGBoost)
            \item Recurrent Neural Networks (with GRU and LSTM as their representatives) 
        \end{itemize}
    
        The rationale for selecting these methods is outlined below. The baseline approach for virtually any time series forecasting task is applying a statistical method and assess its performance. To this end, we exploit on autoregressive methods, training (for each time series) three models from this family, namely ARIMA, SARIMA, and SARIMAX \cite{Box2015Book,Hyndman2021Book}. The primary distinction among these models is that ARIMA is considered baseline, SARIMA explicitly accounts for seasonality, while SARIMAX extends this by incorporating external covariates via a regression component. Specifically, in SARIMAX, regression errors are modeled using SARIMA residuals to capture external influences while retaining seasonal and autoregressive patterns.

        While Transformer-based architectures excel in capturing long-range dependencies, they often require vast datasets to outperform traditional statistical and recurrent models like SARIMAX or GRU, which remain more computationally efficient and robust for localized energy demand forecasting \cite{10.1609/aaai.v37i9.26317}. Given the specific temporal resolution and data volume of this study, the selected models provide a more parsimonious and interpretable balance between complexity and predictive accuracy.

        To select optimal hyperparameters for each model, we leverage AutoML techniques, specifically AutoARIMA\footnote{pmdARIMA: ARIMA estimators for Python. \url{https://alkaline-ml.com/pmdarima/}. Last visited on Sep. 18, 2025.}, and fine-tune each instance to the needs of each time series.

        Another increasingly popular approach for the task at hand, is to use gradient boosted trees \cite{Skaloumpakas2024}, as - compared to SARIMAX - they have been shown to be more capable of capturing complex, non-linear relationships between the input features and the target variable \cite{10.1145/2939672.2939785}, and can handle complex patterns by engineering features representing different seasonal components and/or factors. 

        Moving to deep learning and NNs, RNN-based architectures, such as, the LSTM and GRU \cite{Hochreiter1997,DBLP:conf/ssst/ChoMBB14} are well-known for their potential to handle complex (multi-)variate time series and provide highly accurate results, but at the cost of higher training complexity/cost. 
        While unidirectional models process sequences from past to future, bidirectional architectures (e.g., BiLSTM and BiGRU) integrate future context, which is essential for capturing symmetric dependencies in intermittent demand patterns. By comparing these variants, we aim to evaluate whether the computational overhead of bidirectional processing translates to improved performance in modeling sparse, non-zero demand events.
        
        In brief, the two variants of the RNN model we adopt consist of the following layers: a) an input/recurrent layer for the projection of the time-dependent variable(s), b) a [Bi]LSTM/GRU hidden layer composed of $12$ neurons, c) two embedding layers with $15$ and $3$ neurons for projecting the location and model of the EVSE, respectively, d) a Dropout layer with probability $0.13$ for model regularization, e) a fully-connected hidden layer composed of $12$ neurons, and f) an output layer of $1$ neuron, for the estimated energy demand at the next temporal interval. These hyperparameters were determined through empirical tuning to balance model complexity and generalization, with the output layer size set to correspond to the required number of lookahead steps for demand prediction.
        
        To better direct both XGBoost and RNN models toward non-zero demand periods, we use the Pinball loss function (Equation \ref{eq:PinballLoss}) with $\alpha = 0.7$. According to the literature, values of $\alpha > 0.5$ bias the loss toward under-prediction penalties, thereby encouraging more accurate quantile estimates in the presence of intermittent positive demand \cite{DBLP:conf/acl/VedulaDJOBM25}.
        
        \begin{equation}
            L_{\alpha}(y, \hat{y}) = \dfrac{1}{\lvert y \rvert} \sum_{i=1}^{\lvert y \rvert}{ 
                \max(
                    \alpha (y_i - \hat{y_i}), (\alpha - 1) (y_i - \hat{y_i})
                )
            }
            \label{eq:PinballLoss}
        \end{equation}

        \begin{figure}[!ht]
            \centering
            \includegraphics[width=\columnwidth]{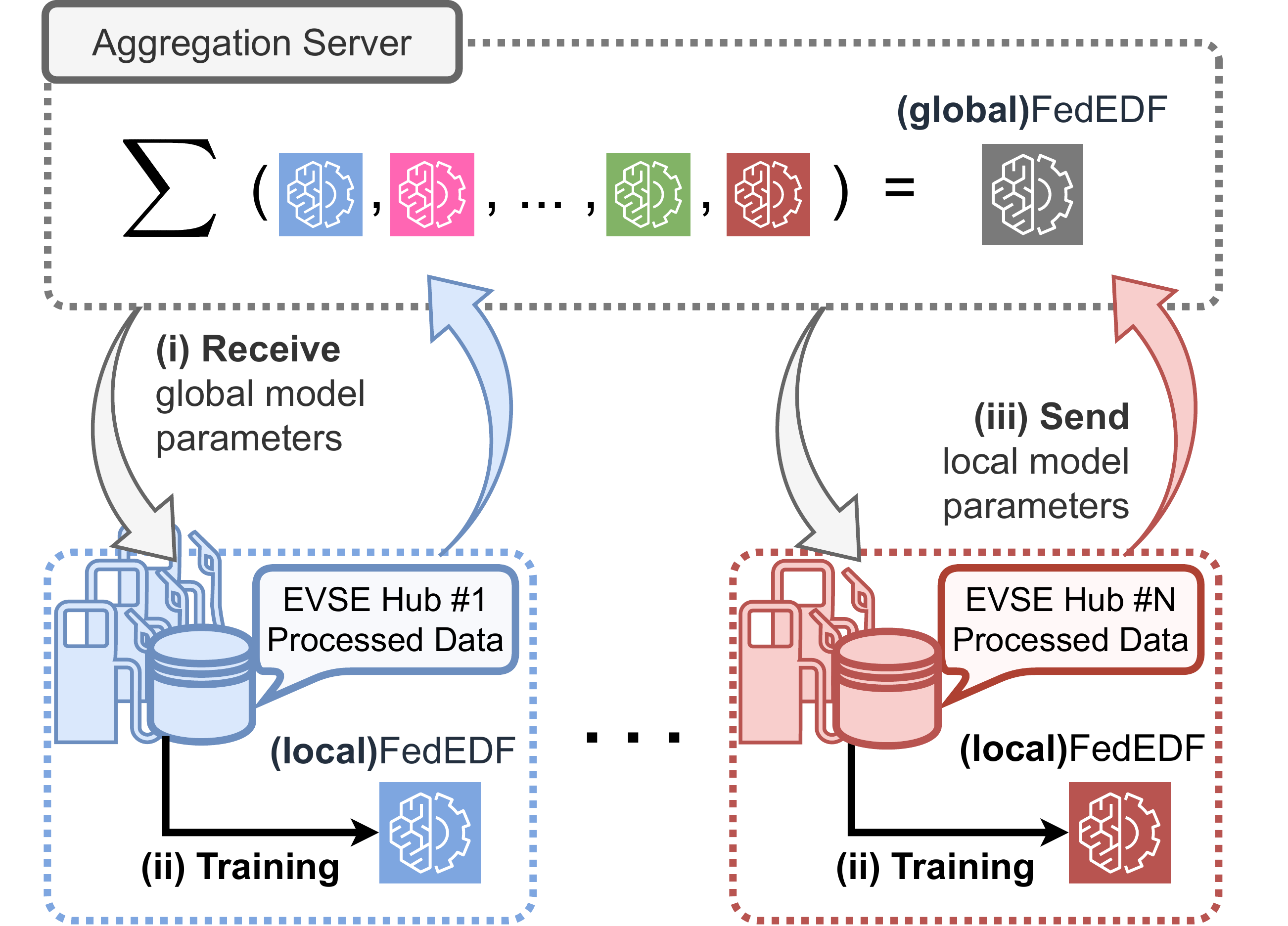}
            \caption{Adapting the EDF experimental comparison workflow for the FL paradigm.}            
            \label{fig:fededf-workflow}
        \end{figure}

    \subsection{Extending to Federated Learning}\label{subsec:MethodologyFL}
        Traditional centralized approaches often face challenges, such as privacy constraints and communication overhead. These issues are particularly significant in real-world EVSE systems, where historical energy demand data is inherently localized to individual entities and subject to privacy regulations. To address these challenges, we experiment with the FL paradigm, which enables collaborative model training across multiple participants without requiring raw data to be shared or centralized. This approach preserves privacy, minimizes communication costs, and potentially improves forecasting accuracy by leveraging the collective insights of distributed clients.

\begin{figure}
            \centering
            \subfloat[\label{subfig:DundeeEVSEDatasetSnap}]{
                \includegraphics[width=0.9\columnwidth]{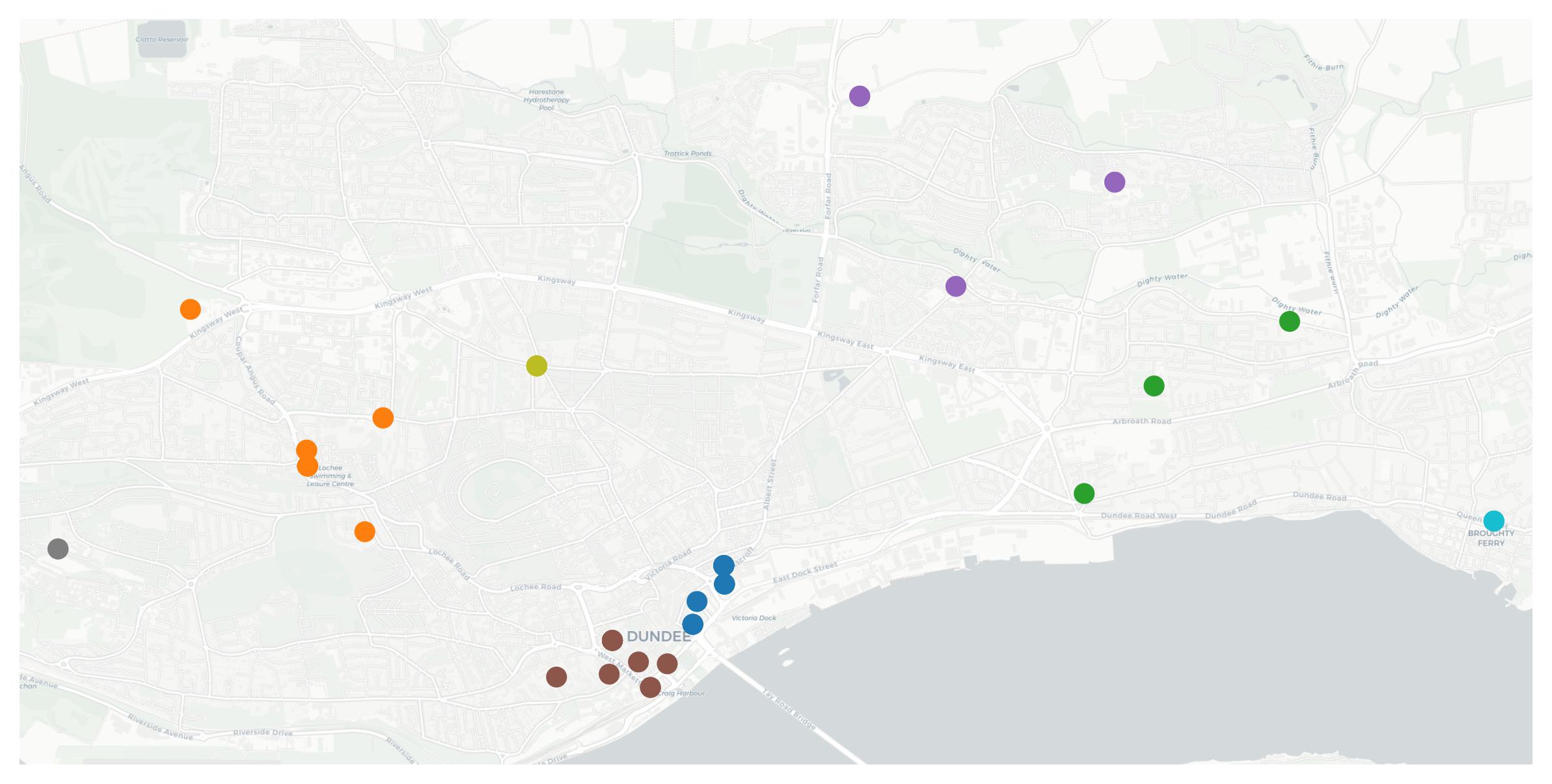}}
            \\
            \subfloat[\label{subfig:FEUPEVSEDatasetSnap}]{
                \includegraphics[width=0.9\columnwidth]{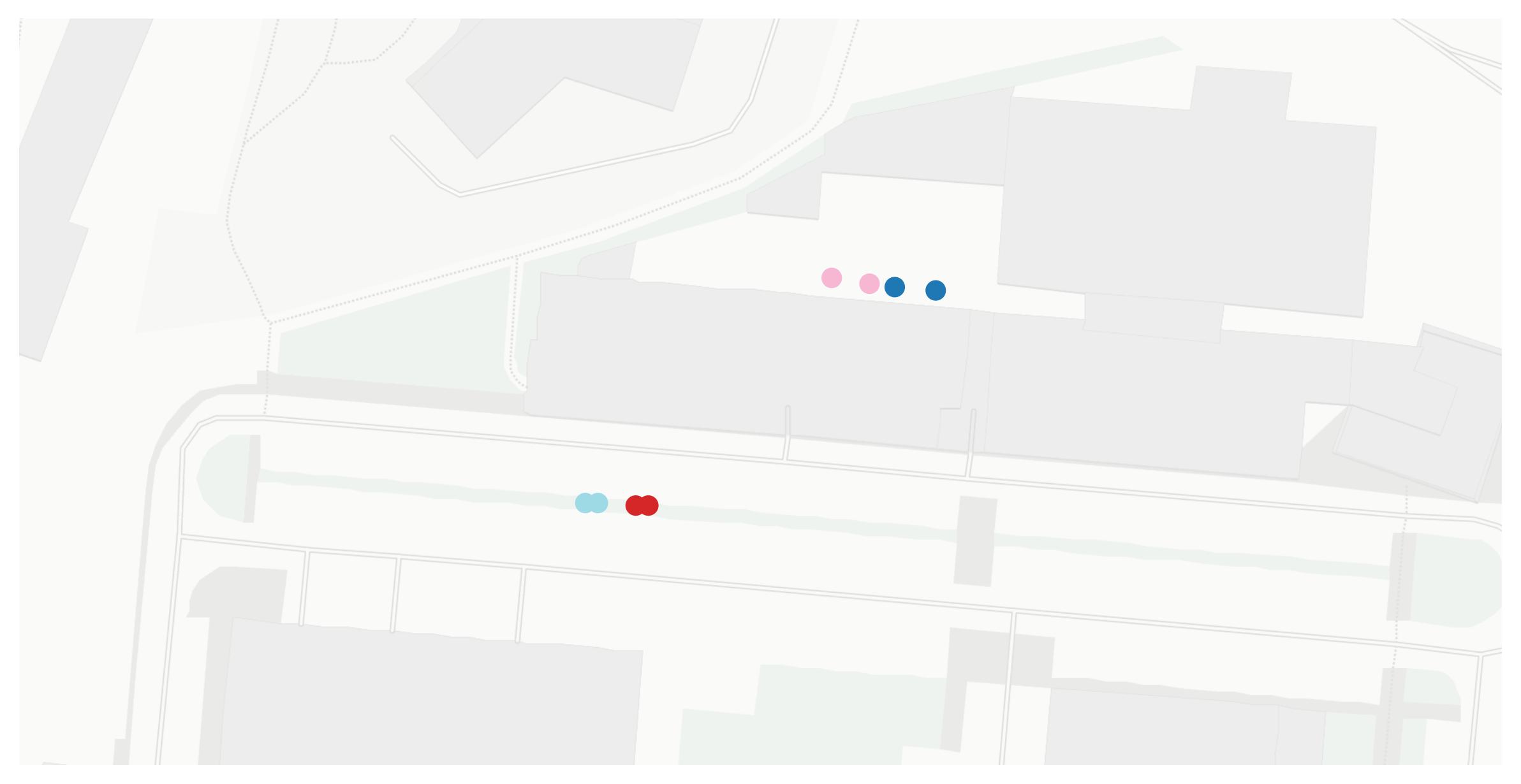}}
            \\
            \subfloat[\label{subfig:BoulderEVSEDatasetSnap}]{
                \includegraphics[width=0.9\columnwidth]{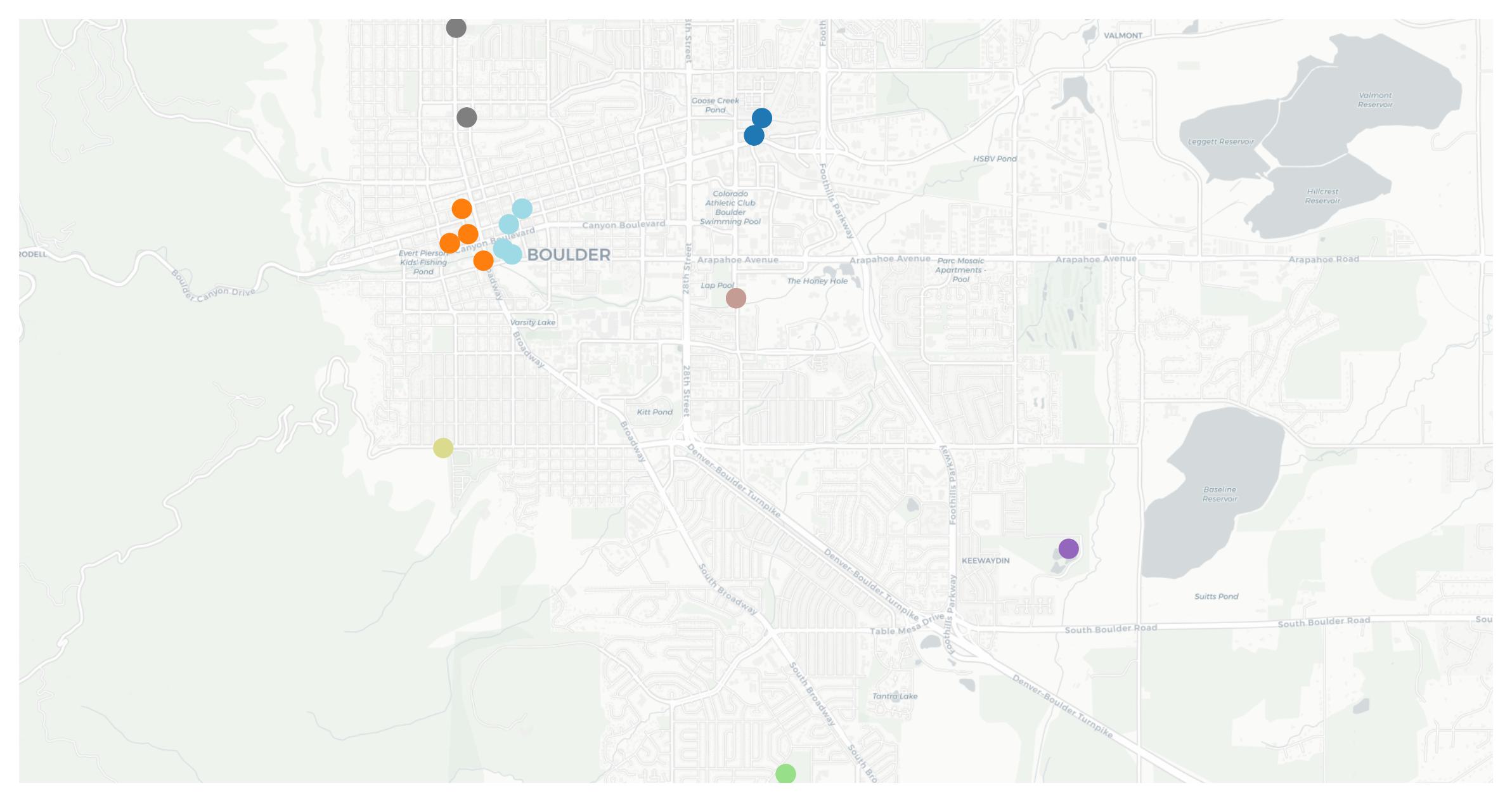}}
            \\
            \subfloat[\label{subfig:PaloaltoEVSEDatasetSnap}]{
                \includegraphics[width=0.9\columnwidth]{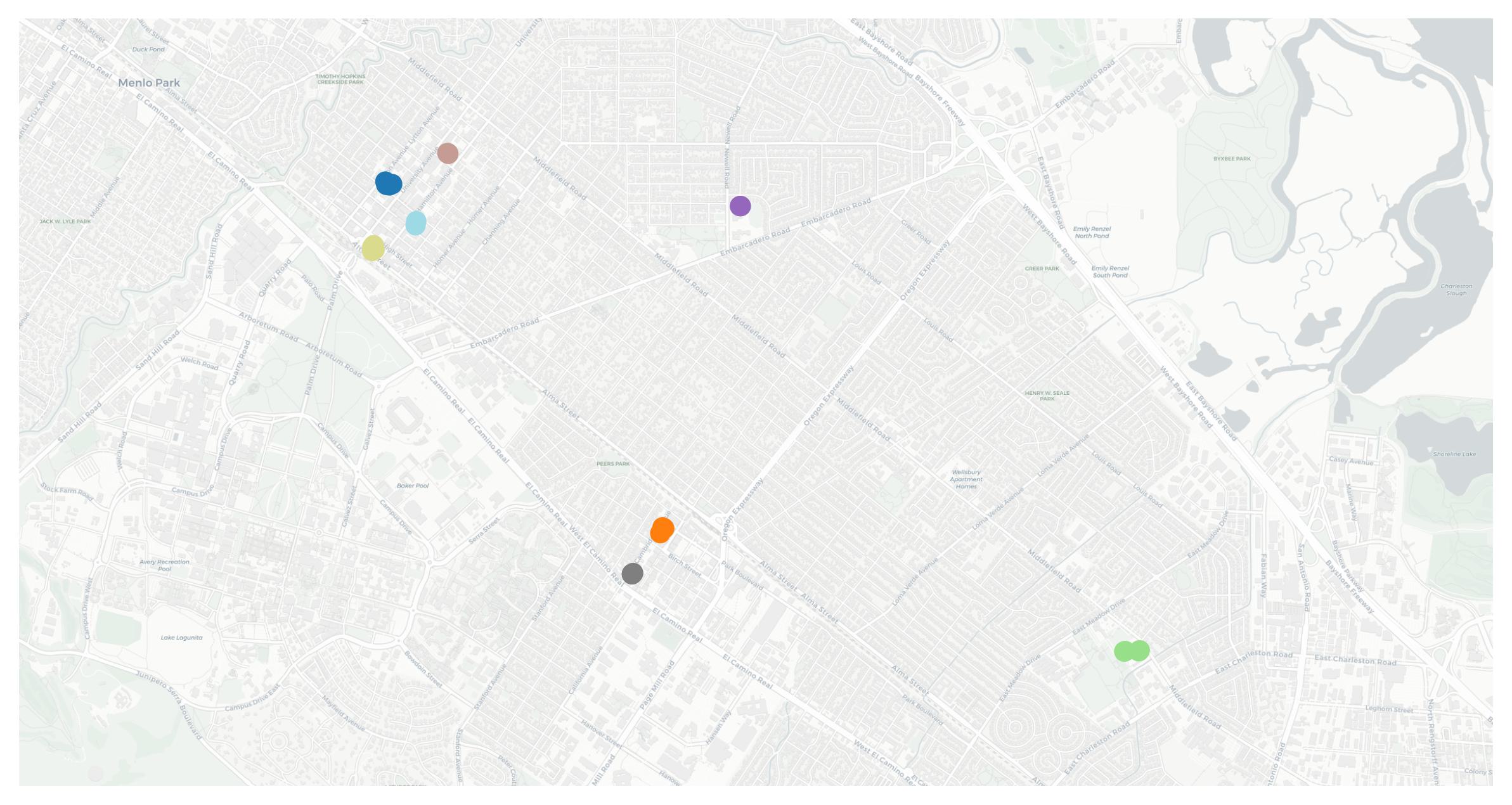}}
            \caption{Snapshot of the EVSE locations of the (a) Dundee; (b) FEUP; (c) Boulder; and (d) Palo Alto dataset. Each color represents a different cluster / EVSE hub.}
            \label{fig:EVSEDatasetSnap}
        \end{figure}

        Figure \ref{fig:fededf-workflow} illustrates our FL-based experimental comparison workflow. We define $N$ clients (i.e., EVSE hubs), each containing aggregated historic energy demand data from a cluster of nearby EVSEs. These EVSE hubs were simulated by running the k-Means algorithm over EVSEs' geographical (lat/lon) location. Combining nearby charging stations alleviates the scarcity of per-station samples, thereby providing the forecaster with a statistically robust dataset without compromising spatial granularity. Figure \ref{fig:EVSEDatasetSnap} illustrates snapshots of EVSE locations and the resulting EVSE hubs on the datasets used in our experimental study, more details of which will be presented in Section \ref{subsec:ExperimentalSetup}.
        
        To better understand how FL works, a round corresponds to a full cycle of global model distribution to clients and subsequent aggregation of updated local models, while an epoch denotes the number of iterations over the clients' local data during model training. At the start of the FL process, the aggregation server randomly selects one participant, and uses its local model as the initial global model. Each FL round involves three sequential steps: (i) sending the parameters of the current global model to the participating clients in the federation, (ii) training the model locally on clients' corresponding data for a set number of local epochs, and (iii) receiving the newly trained local models in order to generate the updated global model.

        In our framework, each client performs $E = 5$ local epochs per round with early stopping patience of $2$ epochs to mitigate overfitting \cite{DBLP:series/lncs/Prechelt12}, over $R = 50$ total rounds. For training the (local)FedEDF models, we adopt the same methodology as the centralized baselines. The (global)FedEDF model is optimized using FedProx \cite{DBLP:conf/mlsys/LiSZSTS20} with $\mu_{prox} = 10^{-1}$. These hyperparameters ($E$, $R$) and $\mu_{prox}$ were empirically selected based on the values listed in \cite{DBLP:conf/mlsys/LiSZSTS20} and validated through convergence analysis of the (local)FedEDF models. 

    \subsection{Summary of Forecasting Methods}\label{subsec:FLCompliance}
% \usepackage{booktabs}
% \usepackage{multirow}
\begin{table*}[!ht]
    \caption{Summary of data-driven methods in the FedEDF framework}    
    \label{tab:model-fl-compliance}
    \centering    
    \renewcommand{\arraystretch}{2.5}
    \resizebox{\textwidth}{!}{%
        \begin{tabular}{@{}lllll@{}}
            \toprule
              Model   
            & Methodology 
            & FL Compatible
            & Hyperparameter Tuning
            & Notes                
            \\ \midrule
              {[}S{]}ARIMA{[}X{]}  
            & Statistical 
            & No            
            & \renewcommand{\arraystretch}{1.1}\begin{tabular}{@{}l@{}}Akaike information criterion\\(AIC)-based order selection\end{tabular}
            & Trained per EVSE; inherently privacy-preserving
            \\
              XGBoost
            & Machine Learning
            & Yes 
            & Empirical, early stopping
            & \renewcommand{\arraystretch}{1.1}\begin{tabular}{@{}l@{}}Trained once for all EVSEs;\\FL implementation aggregates local models via FedXGBllr \cite{DBLP:conf/eurosys/MaQBL23}\end{tabular}
            \\
              {[}Bi{]}\{LSTM, GRU\}   
            & Neural Network   
            & Yes 
            & Empirical, early stopping
            & \renewcommand{\arraystretch}{1.1}\begin{tabular}{@{}l@{}}Trained once for all EVSEs;\\FL implementation aggregates client-side weights via FedProx \cite{DBLP:conf/mlsys/LiSZSTS20}\end{tabular}
            \\ \bottomrule
        \end{tabular}%
    }
\end{table*}
        \begin{table*}[!ht]
    \caption{Statistics of the datasets used in our experimental study, after the preprocessing phase.}
    \label{tab:dataset-stats}
    \centering    
    \renewcommand{\arraystretch}{1.3}
    \resizebox{\textwidth}{!}{%
        \begin{tabular}{@{}llllllll@{}}
            \toprule
                                                & Temporal range        
                                                & \#Transactions
                                                & \#EVSEs
                                                & \renewcommand{\arraystretch}{1.2}\begin{tabular}[c]{@{}l@{}}\#Locations\\ (i.e., distinct sites)\end{tabular} 
                                                & \#EVSE Hubs
                                                & \renewcommand{\arraystretch}{1.2}\begin{tabular}[c]{@{}l@{}}Charging duration (\#minutes)\\ (min.; avg.; max.)\end{tabular}                                   
                                                & \renewcommand{\arraystretch}{1.2}\begin{tabular}[c]{@{}l@{}}Idle Activity (\%)\\ (min.; avg.; max.)\end{tabular}
                                                \\ \midrule
            \rotatebox[origin=c]{0}{Dundee}     & \renewcommand{\arraystretch}{1.2}\begin{tabular}[c]{@{}l@{}}2017-01-09 – \\ 2018-12-05 (696 days)\end{tabular} 
                                                & 47,854        
                                                & 67                                                                   
                                                & 34                                                                         
                                                & 8                                                       
                                                & 1.00; 163.50; 2876.00
                                                & 0.06; 0.65; 0.88
                                                \\
            \rotatebox[origin=c]{0}{FEUP}       & \renewcommand{\arraystretch}{1.2}\begin{tabular}[c]{@{}l@{}}2023-11-13 - \\ 2025-06-20 (586 days)\end{tabular}
                                                & 2,023         
                                                & 12                                                            
                                                & 1                                                                          
                                                & 4                                                 
                                                & 0.08; 372.60; 2476.30
                                                & 0.54; 0.65; 0.80
                                                \\ 
            \rotatebox[origin=c]{0}{Boulder}    & \renewcommand{\arraystretch}{1.2}\begin{tabular}[c]{@{}l@{}}2018-01-02 - \\ 2021-04-01 (1185 days)\end{tabular}
                                                & 21,707
                                                & 27
                                                & 20
                                                & 8
                                                & 1.00; 159.90; 2849.00
                                                & 0.24; 0.60; 0.88
                                                \\ 
            \rotatebox[origin=c]{0}{Palo Alto}  & \renewcommand{\arraystretch}{1.2}\begin{tabular}[c]{@{}l@{}}2011-07-29 - \\ 2021-01-01 (3444 days)\end{tabular}
                                                & 259,308
                                                & 47
                                                & 10
                                                & 8
                                                & 1.00; 149.54; 2822.00
                                                & 0.03; 0.21; 0.76
                                                \\ 
            \bottomrule
        \end{tabular}%
    }
\end{table*}

        Our framework incorporates a diverse set of data-driven time series forecasting methodologies, each selected to align with specific deployment scenarios and privacy constraints. Table \ref{tab:model-fl-compliance} provides a structured summary of these models and their characteristics. In particular, the XGBoost, (Bi)GRU, and (Bi)LSTM models are fully compatible with the Federated Learning (FL) paradigm, enabling collaborative model training across distributed entities without sharing raw data. Hyperparameter tuning was done empirically using each model's performance on the validation set, balancing performance and computational efficiency. On the other hand, ARIMA (and related variants) is inherently decentralized, as it is trained independently on the data of each EVSE, thus eliminating the need for FL. This approach ensures privacy by default, since no data is transmitted between entities.

%% Experimental Study
\section{Experimental Study} \label{sec:experiments}
    In this section, we compare the efficency of the EDF methodologies presented in Section \ref{sec:Background_and_Definitions} using four real-world EVSE energy demand datasets, and present our experimental results. All conducted experiments were implemented in Python. More specifically, the aforementioned models were implemented using PyTorch\footnote{PyTorch: An Imperative Style, High-Performance Deep Learning Library. \url{https://pytorch.org}} and trained using Flower\footnote{Flower: A Friendly Federated Learning Framework. \url{https://flower.ai}} for FL, via a compute node equipped with 256 CPUs, and 1TB of RAM.

    \subsection{Experimental Setup, Datasets and Preprocessing}\label{subsec:ExperimentalSetup}
        For the purposes of our experimental study, we use four real-world EVSE energy demand datasets, including three popular open datasets, hereafter referred to as Dundee, Palo Alto, and Boulder\footnote{The three datasets are publicly available at the GitHub page of \cite{en14082233}.} and one proprietary dataset, provided by FEUP (Faculdade de Engenharia da Universidade do Porto)\footnote{The FEUP dataset is publicly available at: \url{https://doi.org/10.5281/zenodo.17674369}.}. 

        \begin{figure*}[!ht]
            \centering
            \subfloat[\label{subfig:DundeeDatasetSnap}]{
                \includegraphics[width=\columnwidth]{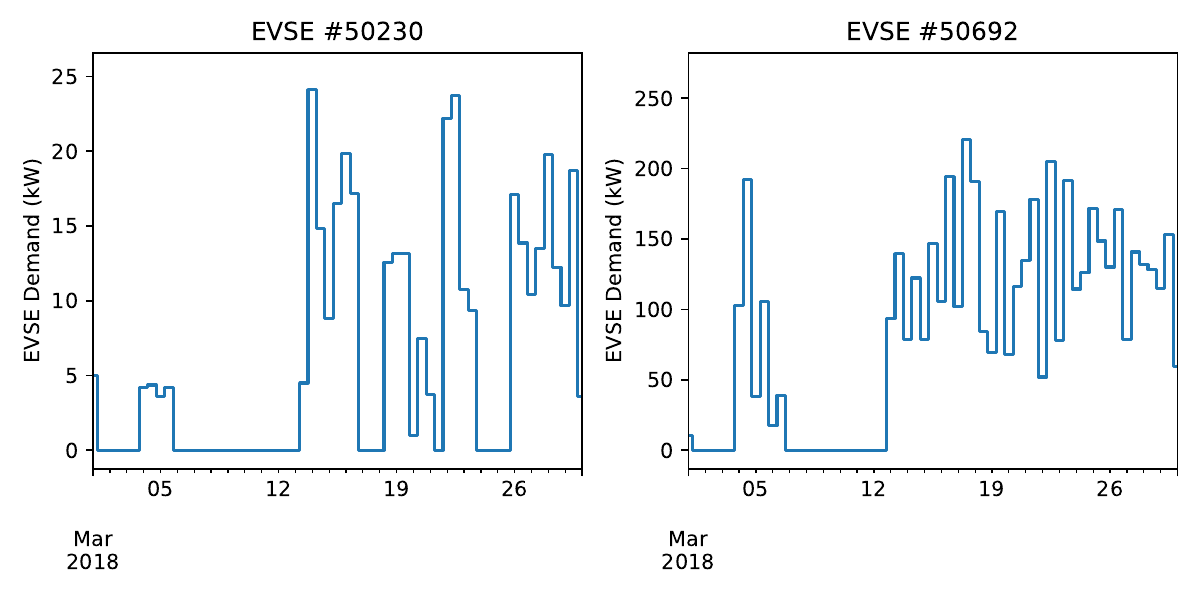}}
            \hfill
            \subfloat[\label{subfig:FEUPDatasetSnap}]{
                \includegraphics[width=\columnwidth]{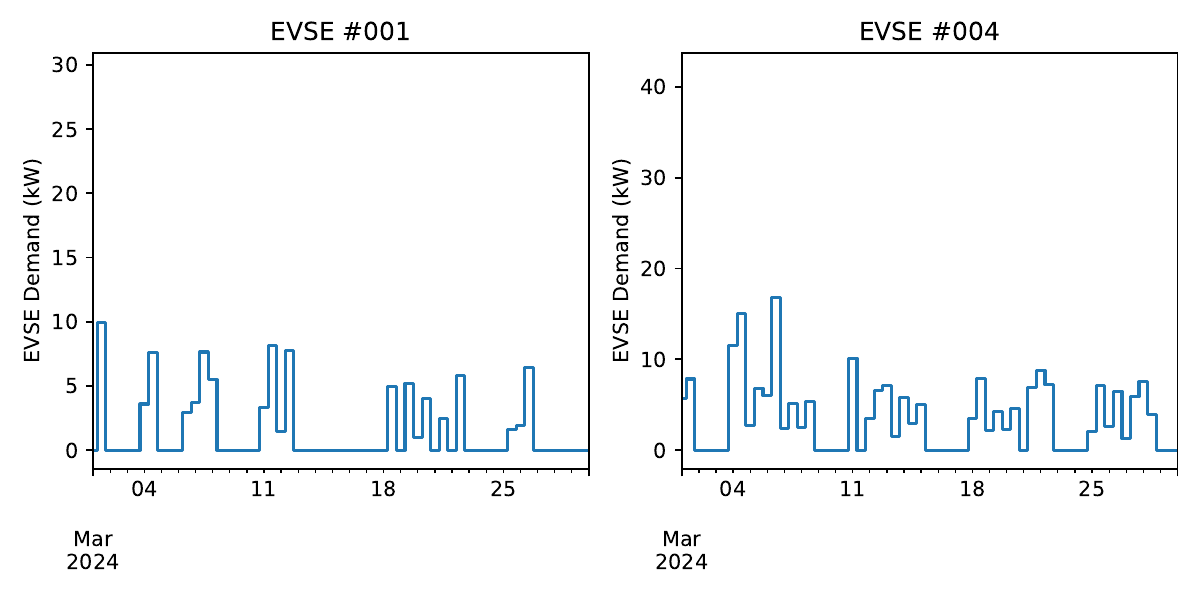}}
            \\
            \subfloat[\label{subfig:BoulderDatasetSnap}]{
                \includegraphics[width=\columnwidth]{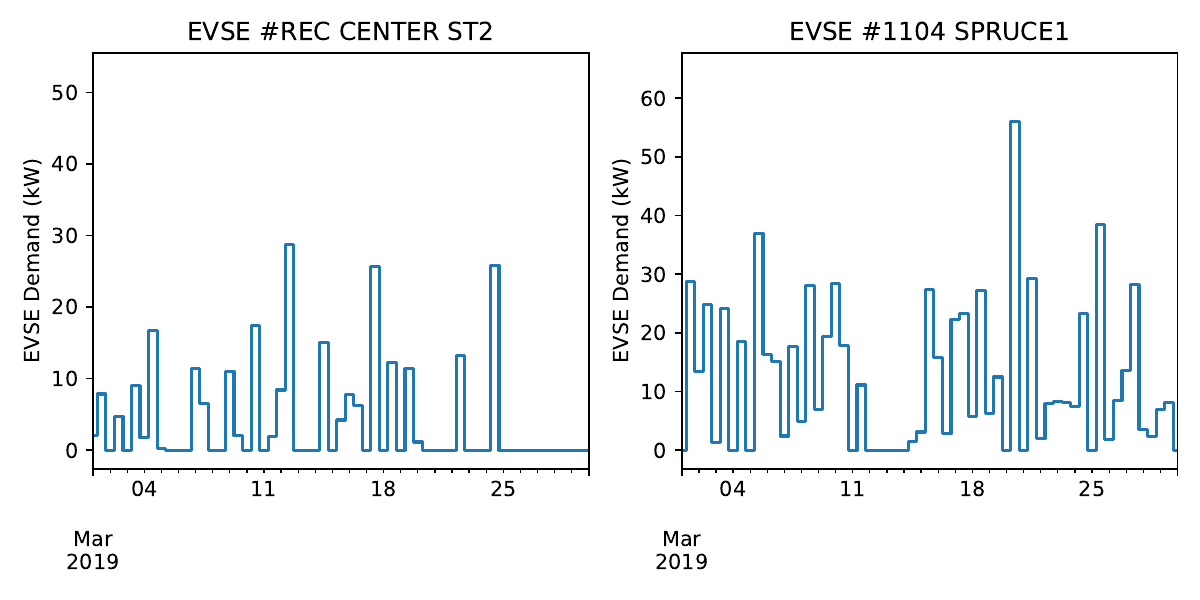}}
            \hfill
            \subfloat[\label{subfig:PaloaltoDatasetSnap}]{
                \includegraphics[width=\columnwidth]{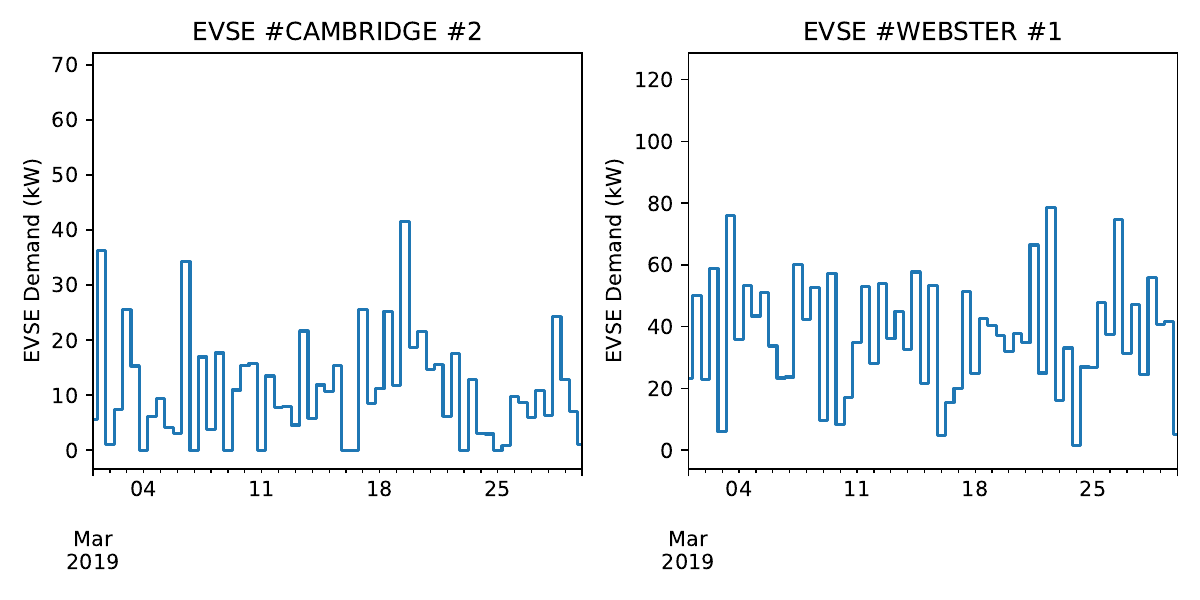}}
            \caption{Snapshots of indicative EVSE energy demand time series (in kW) during March in (a) Dundee; (b) FEUP; (c) Boulder; and (d) Palo Alto dataset, respectively.}
            \label{fig:DatasetSnapshots}
        \end{figure*}
        
        Figure \ref{fig:DatasetSnapshots} provides an illustration of indicative time series included in the four datasets at hand. From this illustration, it is clear that EVSEs' energy demand is intermittent (see, e.g., Dundee EVSE \#50230, Boulder EVSE \#REC CENTER ST2) and (in some cases) lumpy (see, e.g., FEUP EVSE \#004, Palo Alto EVSE \#CAMBRIDGE \#2), thus requiring an effective data preparation approach to assist the model in ``learning'' the EV charging/demand patterns, also considering non-demand periods. 
        For instance, feature engineering incorporates cyclical time features (e.g., hour of day, day of week) to encode temporal context, while sliding windows are used to extract sequential dependencies. To address scale differences across EVSEs, normalization strategies are also required, enabling the model to learn effectively from both active and idle (i.e., non-demand) periods.
        
        To be more precise, our data (pre-)processing stage consists of two phases. The first phase (data cleaning) is a process consistently followed in the literature due to the noise, irregularity of sampling rate, etc. that are typical in time series datasets, whereas the second phase (data preparation for model training) uses the output of the first phase as its input and is performed for ML-specific purposes, i.e., to feed its output into the architectures mentioned in Section \ref{subsec:Methodology}. 

        In particular, the data cleaning phase includes dropping energy transactions $et^{ij}$ (i.e., charging events) with erroneous (e.g., NaN), as well as negative (e.g., Vehicle-to-Grid) consumption values, since our focus within the scope of this paper is solely on Grid-to-Vehicle (G2V) EVs\cite{10883101}. Moreover, we drop outlier transactions with consumed energy $e^{ij}$ greater than $200$ kWh or/and duration higher than $48$ hours, as well as any EVSE $j$ with $ \lvert et^{ij} \rvert < 100$ transactions \cite{en14082233}.

        \begin{figure*}
            \centering
            \subfloat[\label{subfig:ResultsBaselinesDundee}]{
                \includegraphics[width=0.49\columnwidth]{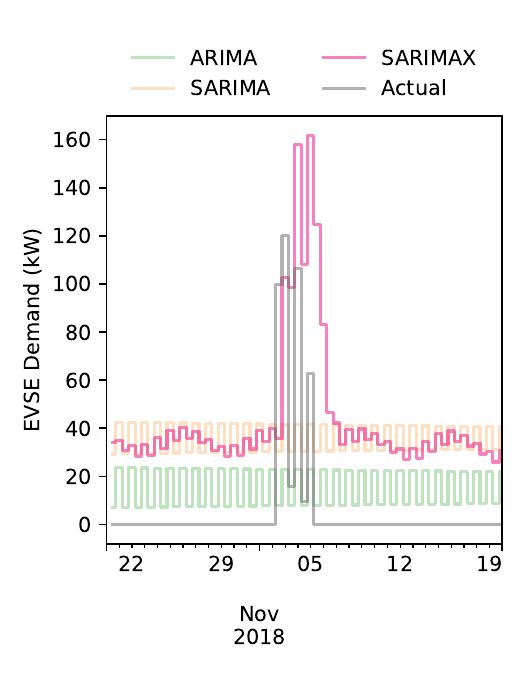}}
            \hfill  
            \subfloat[\label{subfig:ResultsBaselinesFEUP}]{
                \includegraphics[width=0.49\columnwidth]{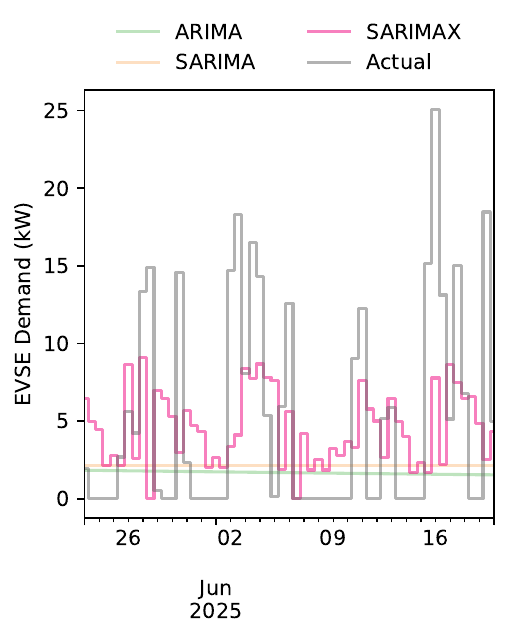}}
            \hfill  
            \subfloat[\label{subfig:ResultsBaselinesBoulder}]{
                \includegraphics[width=0.49\columnwidth]{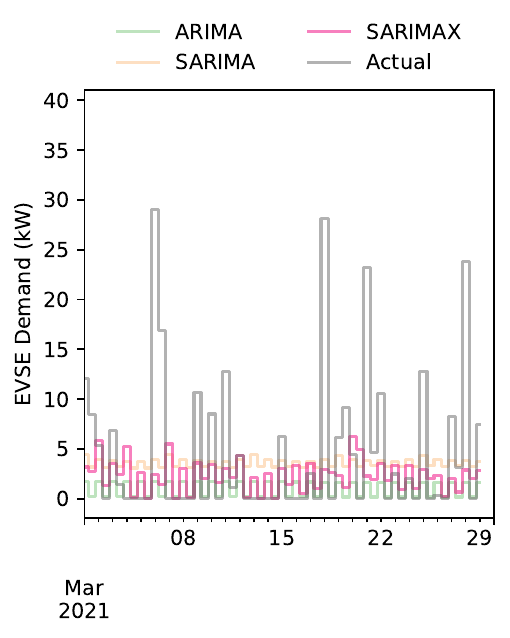}}
            \hfill  
            \subfloat[\label{subfig:ResultsBaselinesPaloAlto}]{
                \includegraphics[width=0.49\columnwidth]{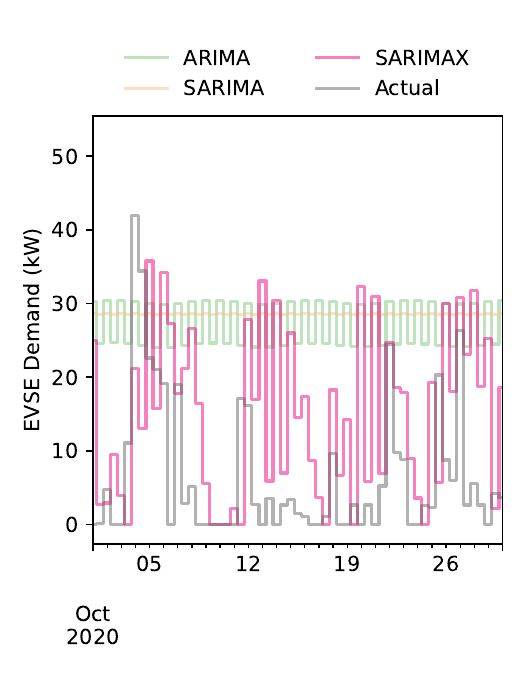}}
            \\
            \subfloat[\label{subfig:ResultsMLDundee}]{
                \includegraphics[width=0.49\columnwidth]{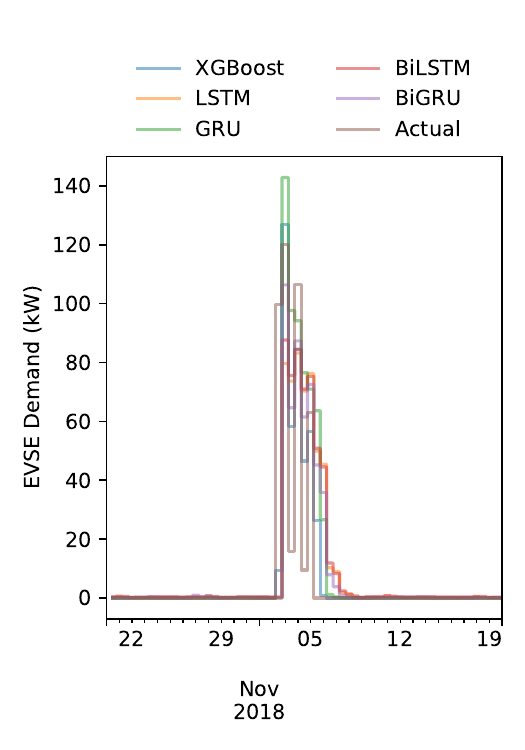}}
            \hfill
            \subfloat[\label{subfig:ResultsMLFEUP}]{
                \includegraphics[width=0.49\columnwidth]{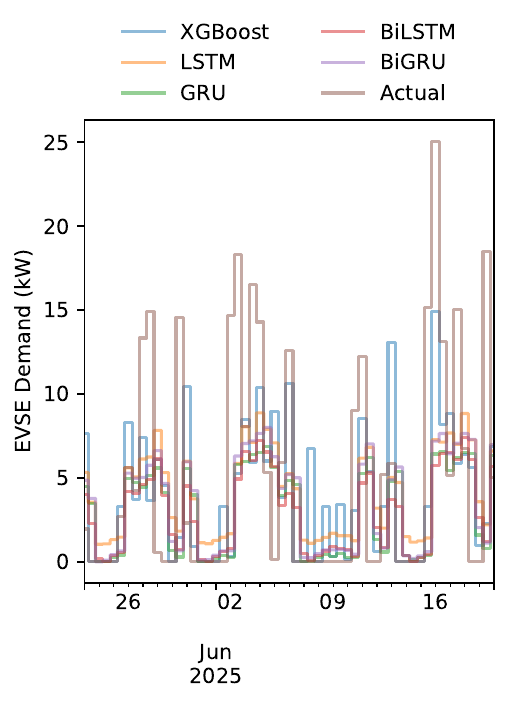}}
            \hfill
            \subfloat[\label{subfig:ResultsMLBoulder}]{
                \includegraphics[width=0.49\columnwidth]{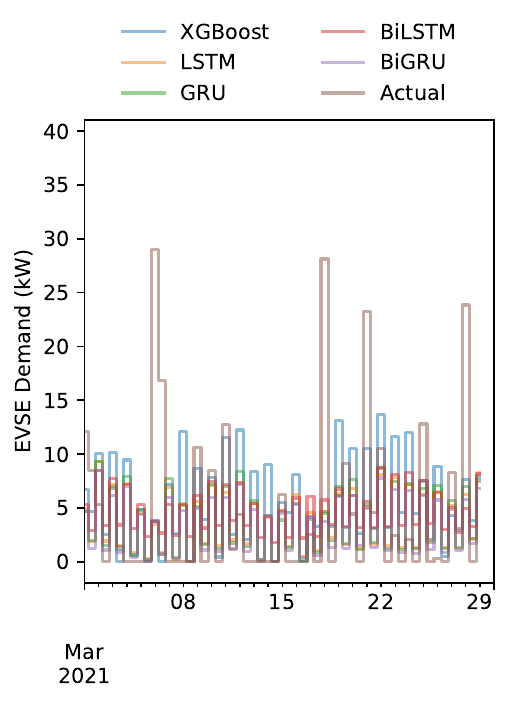}}
            \hfill
            \subfloat[\label{subfig:ResultsMLPaloAlto}]{
                \includegraphics[width=0.49\columnwidth]{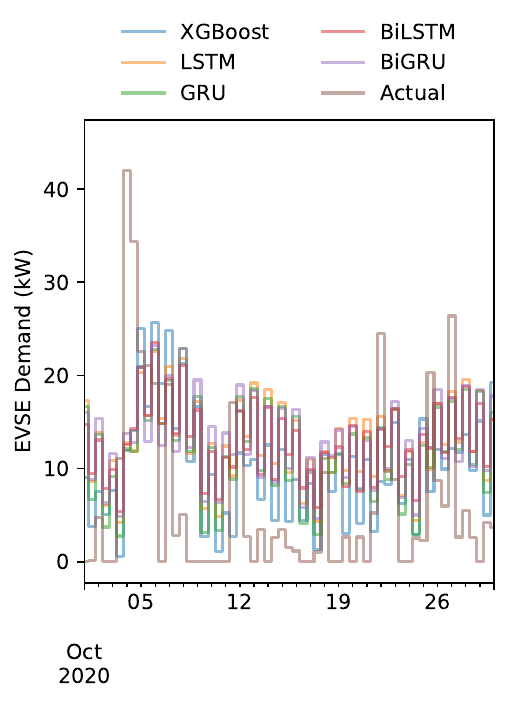}}
            \caption{Energy Demand Forecasting on a randomly selected EVSE from the test set of the Dundee (a, e), FEUP (b, f), Boulder (c, g), and Palo Alto (d, h) Datasets using statistical (a, b, c, d); and ML-based (e, f, g, h) methods, respectively.}
            \label{fig:actual-vs-edf-results}
        \end{figure*}
        
        Then, in the data preparation for model training phase, in order to be compliant with the problem definition presented in Section \ref{subsec:ProblemDefinition}, we convert the cleansed dataset from its original transaction-based format to time series by resampling EVSEs' records to a defined frequency ($sr_{freq} = 12$ hours) based on a common temporal axis (i.e., starting timestamp) with the consumed energy being distributed evenly across the resulting temporal intervals (cf., Definition \ref{def:EnergyDemand}). Table \ref{tab:dataset-stats} illustrates the statistics of the datasets (temporal range, number of locations and EVSE hubs, usage statistics, etc.) after the above preprocessing steps.

        To assess the quality of our models we employ seven well-known metrics, namely, Mean Absolute Scaled Error (MASE), with a 12-day seasonal reference, Symmetric Mean Absolute Percentage Error (SMAPE), Mean Arctangent Absolute Percentage Error (MAAPE), Weighted Absolute Percentage Error (WAPE), Root Mean Squared Error (RMSE), Mean Absolute Error (MAE), and the coefficient of determination (R$^2$). Each metric highlights a different aspect of forecasting performance for the next 12-hour period.

        MASE and $R^2$ evaluate the relative performance of our models against a seasonal baseline, while SMAPE, MAAPE, and WAPE translate errors into percentage terms that are readily interpretable. RMSE and MAE quantify absolute deviations in the unit of measure of the forecast (kW). Each metric emphasizes different aspects of the forecasting error, such as outliers, low-demand periods, or overall variance, so that our evaluation remains reliable across the wide range of operating conditions typical of EVSE charging demand.

    \subsection{Experimental Results, Part I: Centralized Learning}\label{subsec:CentralizedModels}
    
        \begin{table*}[!ht]
    \caption{Prediction error ($25^{th}; 50^{th}; 75^{th}$ quartile) for each methodology per dataset (MASE--MAE - lower is better; $R^2$ - higher is better).}
    \label{tab:edf-results}
    
    \centering    
    \renewcommand{\arraystretch}{1.5}
    \resizebox{\textwidth}{!}{%
        \begin{tabular}{@{}lllllllll@{}}
            \toprule
            & \multirow{3}{*}{Model} 
            & \multicolumn{7}{c}{Metric}
            \\ \cmidrule(l){3-9} 
            & ~
            & MASE (\%)
            & SMAPE (\%)
            & MAAPE (rads) 
            & WAPE (\%)
            & RMSE (kW) 
            & MAE (kW) 
            & $R^2$
            \\ \midrule
            \multirow{8}{*}{\rotatebox[origin=c]{90}{Dundee}}
            & ARIMA 
            & 0.55; 0.68; 0.79 
            & 1.88; 1.93; 1.95 
            & 1.47; 1.49; 1.51 
            & 2.81; 3.37; 5.19 
            & 1.73; 2.92; 5.81 
            & 1.18; 2.11; 4.63 
            & -0.55; -0.25; -0.10
            \\
            & SARIMA
            & 0.62; 0.73; 0.82 
            & 1.90; 1.93; 1.95 
            & 1.47; 1.50; 1.51 
            & 2.86; 3.59; 5.36 
            & 1.77; 2.93; 5.77 
            & 1.37; 2.15; 4.36 
            & -0.61; -0.28; -0.16
            \\
            & SARIMAX 
            & 0.36; 0.42; 0.47 
            & 1.04; 1.54; 1.88 
            & 0.80; 1.21; 1.48 
            & 1.32; 2.00; 3.48 
            & 1.58; 2.47; 4.94 
            & 0.76; 1.24; 2.02 
            & -0.44; 0.17; 0.36
            \\        
            & XGBoost
            & 0.14; \textbf{0.17; 0.25} 
            & \textbf{0.14; 0.16; 0.18} 
            & \textbf{0.10; 0.12; 0.13} 
            & \textbf{0.76}; 1.08; \textbf{1.26}
            & 1.28; \textbf{2.37; 4.52}
            & 0.31; \textbf{0.63; 1.17}
            & -0.17; \textbf{0.24}; 0.42
            \\
            & LSTM        
            & 0.14; 0.21; 0.31 
            & 1.06; 1.50; 1.82 
            & 0.82; 1.17; 1.42 
            & 0.86; 1.21; 1.53 
            & \textbf{1.26}; 2.50; 5.27 
            & 0.38; 0.77; 1.41 
            & \textbf{-0.12}; 0.16; 0.41
            \\
            & GRU
            & 0.14; 0.19; 0.27 
            & 0.47; 1.55; 1.84 
            & 0.36; 1.21; 1.44 
            & 0.78; \textbf{1.08}; 1.33 
            & 1.36; 2.50; 5.01 
            & \textbf{0.30}; 0.70; 1.34 
            & -0.29; 0.10; 0.43
            \\
            & BiLSTM
            & 0.15; 0.23; 0.35 
            & 1.82; 1.85; 1.89 
            & 1.42; 1.44; 1.49 
            & 0.99; 1.33; 1.90
            & 1.32; 2.52; 5.41 
            & 0.42; 0.84; 1.93 
            & -0.25; 0.13; 0.40
            \\
            & BiGRU             
            & \textbf{0.12}; 0.21; 0.30
            & 1.67; 1.80; 1.88 
            & 1.30; 1.41; 1.46 
            & 0.85; 1.18; 1.53 
            & 1.30; 2.59; 5.34 
            & 0.37; 0.75; 1.41 
            & -0.20; 0.17; \textbf{0.45}
            \\ \bottomrule
            
            \multirow{8}{*}{\rotatebox[origin=c]{90}{FEUP}}
            & ARIMA 
            & 0.76; 1.00; 1.11 
            & 1.52; 1.59; 1.66 
            & 1.12; 1.15; 1.21 
            & 1.08; 1.12; 1.37 
            & 3.28; 4.36; 6.73 
            & 2.52; 3.04; 4.67 
            & -0.06; -0.02; -0.01
            \\
            & SARIMA 
            & 0.78; 0.99; 1.10 
            & 1.52; 1.59; 1.65 
            & 1.13; 1.14; 1.22 
            & 1.09; 1.11; 1.32 
            & 3.23; 4.41; 6.76 
            & 2.50; 3.10; 4.67 
            & -0.06; -0.04; -0.03
            \\
            & SARIMAX 
            & 0.70; 0.78; 0.92
            & 1.30; 1.45; 1.50
            & 0.94; 1.07; 1.12 
            & 0.90; 0.98; 1.24 
            & 2.96; 4.18; 6.54 
            & 1.91; 2.76; 4.16 
            & -0.02; 0.03; 0.15
            \\
            & XGBoost 
            & \textbf{0.59; 0.69; 0.90}
            & \textbf{0.72; 0.89; 0.94}
            & \textbf{0.56; 0.66; 0.68}
            & \textbf{0.79}; 0.88; 1.13 
            & \textbf{2.87; 4.18}; 6.35 
            & \textbf{1.68; 2.49; 3.81}
            & -0.02; \textbf{0.08; 0.18}
            \\
            & LSTM
            & 0.82; 1.03; 1.10
            & 1.39; 1.45; 1.55 
            & 1.09; 1.12; 1.17 
            & 0.94; 1.03; 1.22 
            & 3.19; 4.92; \textbf{6.20}
            & 2.47; 3.29; 4.39 
            & -0.04; 0.06; 0.14
            \\
            & GRU
            & 0.67; 0.85; 1.00
            & 1.28; 1.31; 1.44 
            & 0.94; 0.97; 1.04 
            & 0.84; 0.90; 1.06 
            & 3.20; 4.97; 6.61 
            & 2.06; 2.84; 4.23 
            & -0.04; 0.02; 0.08
            \\
            & BiLSTM
            & 0.66; 0.83; 0.97 
            & 1.25; 1.32; 1.39 
            & 0.93; 0.96; 1.03 
            & 0.83; \textbf{0.88; 1.01}
            & 2.96; 4.95; 6.52 
            & 2.04; 2.74; 4.17 
            & \textbf{0.00}; 0.08; 0.12
            \\
            & BiGRU          
            & 0.71; 0.86; 1.02 
            & 1.40; 1.42; 1.52 
            & 1.05; 1.06; 1.12 
            & 0.85; 0.92; 1.10
            & 3.24; 4.97; 6.46 
            & 2.15; 2.95; 4.21 
            & -0.02; 0.05; 0.10
            \\ \bottomrule
            
            \multirow{8}{*}{\rotatebox[origin=c]{90}{Boulder}}
            & ARIMA 
            & 0.39; 0.60; 0.72 
            & 1.63; 1.78; 1.91 
            & 1.22; 1.35; 1.45 
            & \textbf{1.12}; 1.80; 2.22 
            & 4.60; 5.95; 7.41 
            & 2.95; 3.31; \textbf{4.15} 
            & -0.09; -0.04; 0.01
            \\
            & SARIMA 
            & 0.54; 0.70; 0.77 
            & 1.54; 1.70; 1.87 
            & 1.21; 1.34; 1.45 
            & 1.57; 2.00; 2.60 
            & 4.60; 5.98; 7.41 
            & 2.95; 3.84; 5.46 
            & -0.13; -0.07; 0.01
            \\
            & SARIMAX 
            & 0.60; 0.68; 0.72 
            & 1.47; 1.65; 1.81 
            & 1.15; 1.26; 1.42 
            & 1.50; 2.06; 2.83 
            & 4.70; 6.50; 7.70
            & 3.21; 3.86; 5.72 
            & -0.42; -0.06; 0.01
            \\
            & XGBoost 
            & \textbf{0.33; 0.43}; 0.61 
            & \textbf{0.77; 1.07; 1.17} 
            & \textbf{0.56; 0.79; 0.89} 
            & 1.24; \textbf{1.38; 1.55}
            & 4.69; \textbf{5.63}; 7.47
            & \textbf{1.90; 2.70;} 4.16
            & -0.13; -0.07; 0.05
            \\
            & LSTM 
            & 0.41; 0.49; 0.60
            & 1.24; 1.40; 1.63 
            & 0.93; 1.07; 1.26 
            & 1.32; 1.40; 1.63 
            & 4.36; 5.84; \textbf{6.80}
            & 2.37; 3.00; 4.45 
            & \textbf{-0.02; 0.03;} 0.07
            \\
            & GRU 
            & 0.39; 0.48; 0.60 
            & 1.18; 1.33; 1.51 
            & 0.91; 1.04; 1.14 
            & 1.20; 1.42; 1.59 
            & 4.30; 5.96; 6.83 
            & 2.27; 2.97; 4.63 
            & -0.05; 0.02; \textbf{0.07}
            \\
            & BiLSTM 
            & 0.62; 0.67; 0.85 
            & 1.60; 1.72; 1.84 
            & 1.25; 1.33; 1.44 
            & 1.69; 1.94; 2.39 
            & 4.67; 6.37; 7.42 
            & 3.39; 3.91; 5.74 
            & -0.32; -0.14; 0.05
            \\
            & BiGRU 
            & 0.42; 0.48; \textbf{0.60}
            & 1.37; 1.47; 1.61 
            & 1.07; 1.13; 1.25 
            & 1.30; 1.46; 1.61 
            & \textbf{4.32}; 5.99; 6.89 
            & 2.28; 2.93; 4.70 
            & -0.07; 0.01; 0.07
            \\ \bottomrule
            
            \multirow{8}{*}{\rotatebox[origin=c]{90}{Palo Alto}}
            & ARIMA 
            & 0.82; 0.95; 1.14 
            & 1.16; 1.34; 1.47 
            & 0.96; 1.11; 1.22 
            & 1.09; 1.58; 2.27 
            & 10.46; 14.70; 18.94 
            & 8.79; 10.58; 16.83 
            & -0.86; -0.46; -0.11
            \\
            & SARIMA 
            & 0.99; 1.14; 1.26 
            & 1.19; 1.38; 1.50 
            & 1.03; 1.16; 1.24 
            & 1.40; 1.82; 2.88 
            & 12.07; 14.71; 19.83 
            & 10.27; 13.12; 17.83 
            & -1.68; -0.98; -0.28
            \\
            & SARIMAX 
            & 0.81; 0.92; 1.05 
            & 1.11; 1.27; 1.38 
            & \textbf{0.90;} 1.04; 1.13 
            & 1.19; 1.51; 1.93 
            & 9.77; 13.16; 17.46 
            & 8.00; 11.23; 14.08 
            & -0.78; -0.40; -0.12
            \\
            & XGBoost 
            & 0.40; \textbf{0.58; 0.69} 
            & \textbf{0.96; 1.14; 1.25} 
            & 0.78; \textbf{0.93; 1.00} 
            & \textbf{0.86; 0.90; 1.08} 
            & 8.29; 9.92; 11.82 
            & 5.90; 7.49; 9.18 
            & 0.03; \textbf{0.29; 0.41}
            \\
            & LSTM 
            & 0.38; 0.59; 0.74 
            & 1.13; 1.32; 1.39 
            & 0.94; 1.06; 1.10 
            & 0.92; 1.11; 1.25 
            & 7.77; \textbf{9.50; 11.24} 
            & 5.61; \textbf{7.35; 8.68} 
            & \textbf{0.04}; 0.14; 0.31
            \\
            & GRU 
            & 0.41; 0.61; 0.73 
            & 1.14; 1.31; 1.46 
            & 0.95; 1.06; 1.16 
            & 0.92; 1.15; 1.37 
            & 7.92; 9.74; 11.48 
            & 5.88; 7.58; 8.95 
            & -0.04; 0.15; 0.26
            \\
            & BiLSTM 
            & 0.45; 0.62; 0.73 
            & 1.10; 1.32; 1.41 
            & 0.92; 1.05; 1.13 
            & 0.94; 1.17; 1.49 
            & 8.08; 9.83; 11.64 
            & 6.23; 7.51; 9.17 
            & -0.13; 0.10; 0.23
            \\
            & BiGRU 
            & \textbf{0.37;} 0.61; 0.76 
            & 1.11; 1.33; 1.38 
            & 0.92; 1.05; 1.12 
            & 0.94; 1.12; 1.30
            & \textbf{7.76;} 9.66; 11.80 
            & \textbf{5.57;} 7.37; 9.33 
            & -0.09; 0.17; 0.29
            \\ \bottomrule
        \end{tabular}
    }
\end{table*}

        % SARIMAX
        As mentioned in Section \ref{subsec:Methodology}, we train one instance of ARIMA and its related variants SARIMA and SARIMAX on each time series using AutoML approaches, in order to select the best model for each EVSE.  
        
        After the preprocessing steps described in Section \ref{subsec:ExperimentalSetup}, we split the temporal axis of our data into training, validation, and test sets using a ratio of 70:20:10\% (1 fold), and train an ARIMA-based model for each time series of the training set. Specifically, the exogenous variables included in the SARIMAX model are the cyclical encodings of hour, day, and week, along with the EVSE activity score (cf. Equation~\ref{eq:ActivityScore}), the number of sessions, and the average charging time per interval.

        \begin{equation}
            \text{activity\_score}_i = e^{-\text{downtime}_i},
            \label{eq:ActivityScore}
        \end{equation}
        where $\text{downtime}_i$ corresponds to the number of temporal intervals since non-zero demand at time step $i$.
        
        Figure \ref{subfig:ResultsBaselinesDundee} illustrates the energy demand forecast for a randomly selected EVSE from the test set of the Dundee dataset, whereas the complete results, in terms of prediction error metrics, are listed in the respective rows of Table \ref{tab:edf-results}. Statistical models (ARIMA, SARIMA, SARIMAX) demonstrate poor accuracy, with ARIMA yielding the second highest MASE, reflecting large relative errors compared to naive forecasts. SARIMA performed worse due to its fixed seasonal structure, while SARIMAX significantly improved accuracy up to $43 \%$, in terms of MASE, by incorporating exogenous variables, lowering MAE to 0.76 -- 2.02 kW, a factor up to $56.37 \%$. However, the high SMAPE scores, suggest that SARIMAX fails to ``capture'' dynamic demand patterns, causing delays in power delivery.

        % XGBoost
        In the next set of experiments, we employ eXtreme Gradient Boosted Trees (XGBoost) to train a single model across all EVSEs in the dataset. In terms of preprocessing, we follow the same steps as before, but instead of a single EVSE, we include all EVSE in the training set. As mentioned in Section \ref{subsec:Methodology}, to model non-linear relationships between the input and target variables, we perform feature engineering. This includes generating synthetic features derived from energy consumption and timestamps. More specifically, we incorporate the same exogenous variables as SARIMAX (excluding the EVSE activity score), along with the $0^{th}$ (current), $1^{st}$, $5^{th}$, and $48^{th}$ lags (i.e., previous measurements) of energy demand, alongside the $0^{th}$, $24^{th}$, and $48^{th}$ lag of rolling standard deviation and exponential moving average, computed over windows of $4$ -- $6$ time steps \cite{Hyndman2021Book,Skaloumpakas2024}. 
        
        To enhance model performance, we incorporate the logarithmic difference between consecutive energy demand values (cf. Equation \ref{eq:EnergyDemandLogDelta}) and an estimation of future energy demand via linear extrapolation. To account for variability in EVSE characteristics (e.g., power output and consumption patterns), we engineer two additional features, namely, nominal power output (in kW), and downtime (i.e., intervals since the last non-zero demand). The nominal power output not only normalizes EVSE demand (as previously described) but also provides insights into charging dynamics: higher power outputs correlate with shorter charging times and narrower charging plateaus.
        \begin{equation}
            \Delta \log P_{j(k,\ k-1)} = \log{(P_{jk} + \epsilon)} - \log{(P_{j(k-1)} + \epsilon)},
            \label{eq:EnergyDemandLogDelta}
        \end{equation}
        where $\epsilon = 10^{-9}$. For normalization, all power-based features are scaled by the maximum power output of each EVSE at each time interval. Charging duration is scaled by the maximum allowed duration (48 hours; cf. Section \ref{subsec:ExperimentalSetup}). Since the values for ``downtime'' and number of sessions are unbounded, in order to prevent data leakage, we normalize them by dividing with the maximum value observed in the train set.
                
        Finally we set as target EVSEs' power demand at the next time step, and proceed to train an XGBoost instance with $50$ estimators, each with a tree of maximum depth at $5$ levels, using the Pinball loss function (cf., Section \ref{subsec:Methodology}) with learning rate $\eta = 10^{-1}$. All hyperparameters were empirically selected through extensive experimentation in order to optimize performance while considering model complexity and generalization.

        Figure \ref{subfig:ResultsMLDundee} illustrates the energy demand forecast for a randomly selected EVSE from the test set of the Dundee dataset, whereas the complete results, in terms of prediction error metrics, are listed in the respective rows of Table \ref{tab:edf-results}. 

        It can clearly be observed that XGBoost outperforms all other models, achieving the lowest MASE score, reflecting higher fidelity in capturing non-linear patterns and interactions between input features. Moreover, its low SMAPE score further highlights minimal symmetric errors, i.e., delay in power delivery, while the increased $R^2$ indicates better alignment with observed trends. These conclusions are also reflected in terms of MAE, which XGBoost reduces to 0.31 -- 1.17 kW, a factor up to $77.37\%$.

        % BiLSTM
        In another set of experiments, we employ RNN-based models, specifically unidirectional LSTM/GRU and bidirectional variants, and train a single model across all EVSEs in the dataset. Our preprocessing workflow is similar to the previous experiments, but instead of a single EVSE, we include all EVSE in the training set. All preprocessing steps, including feature engineering and normalization, follow the exact process described in the XGBoost-related experiments.

        RNN models inherently capture sequential dependencies in the data, eliminating the need to include lagged variables as explicit input features. Given the use of up to the $48^{th}$ lag, we segment EVSEs' time series into fixed-length sequences of $48$ observations using a rolling window with a stride of 1 observation. To efficiently handle categorical variables, we include EVSEs' model and location using separate embedding layers, which are merged with the output of the RNN at the fully connected layer, along with their corresponding nominal power output in (kW).

        Finally we set as target EVSEs' power demand at the next time step, and proceed to train each RNN model using the Adam \cite{DBLP:journals/corr/KingmaB14} optimisation algorithm for 100 epochs with batch size set to 32 training samples. To prevent over-fitting, we use the well-known early stopping \cite{DBLP:series/lncs/Prechelt12} mechanism with a patience of 10 epochs. 

        After training, Figure \ref{subfig:ResultsMLDundee} illustrates the performance of [Bi]LSTM/GRU over the same EVSE used as example the aforementioned experiments, whereas the completed results are listed in the respective rows in Table \ref{tab:edf-results}. It can be observed that the GRU outperforms not only the statistical models, up to $77.42\%$, in terms of MASE, but also all other RNN variants up to $30.56\%$, in terms of MAE. However, the XGBoost remains the better model across all metrics at the 50\textsuperscript{th} percentile, except WAPE, which GRU improves by $0.09\%$. Overall, the high SMAPE scores suggest limitations in adapting to sudden demand shifts (i.e., power delivery delays). Additionally, the broader range in the $R^2$ scores suggest a lower fidelity in estimating future energy demand, potentially caused by the intermittent and lumpy nature of the time series.        

        \begin{figure*}[!ht]
            \centering
            \subfloat[\label{subfig:ResultsFedMLDundee}]{
                \includegraphics[width=0.49\columnwidth]{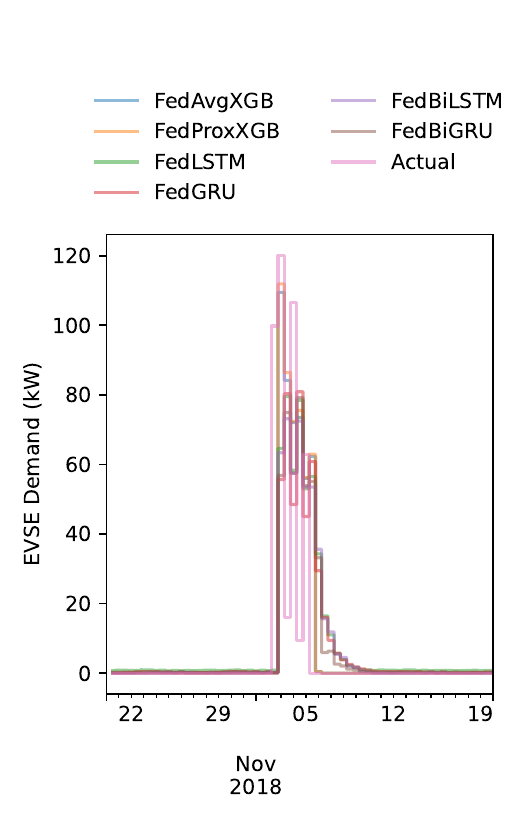}}
            \hfill  
            \subfloat[\label{subfig:ResultsFedMLFEUP}]{
                \includegraphics[width=0.49\columnwidth]{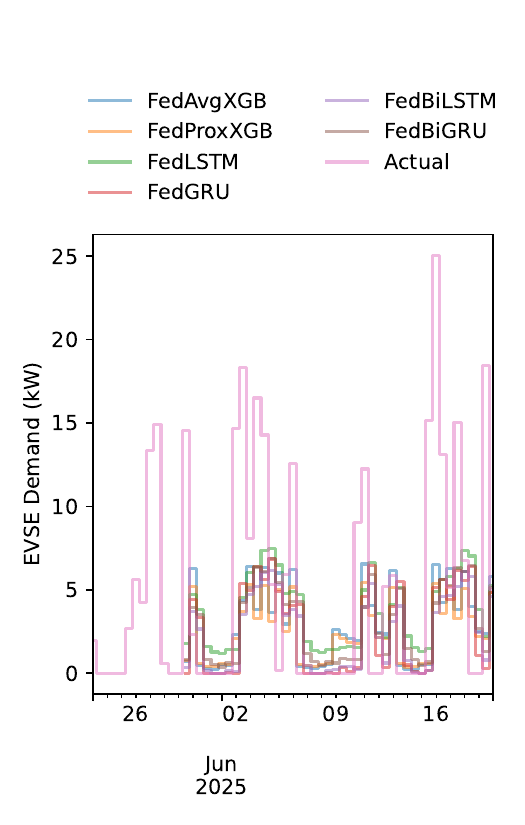}}
            \hfill  
            \subfloat[\label{subfig:ResultsFedMLBoulder}]{
                \includegraphics[width=0.49\columnwidth]{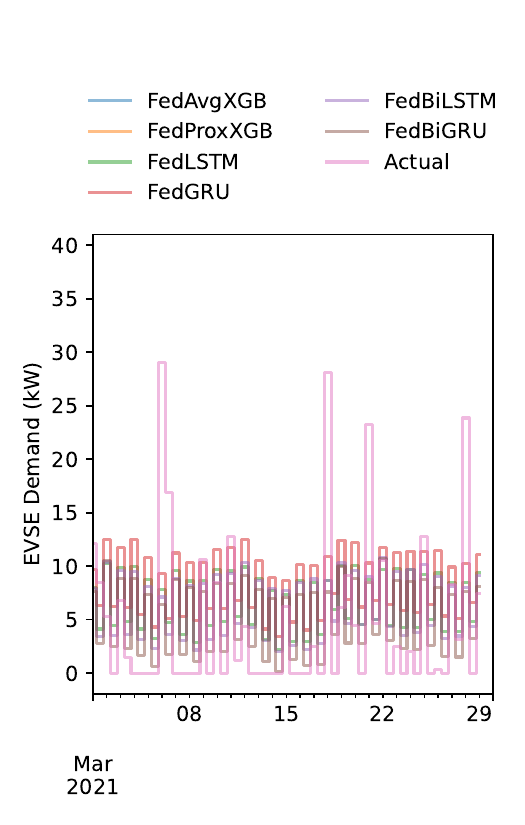}}
            \hfill  
            \subfloat[\label{subfig:ResultsFedMLPaloAlto}]{
                \includegraphics[width=0.49\columnwidth]{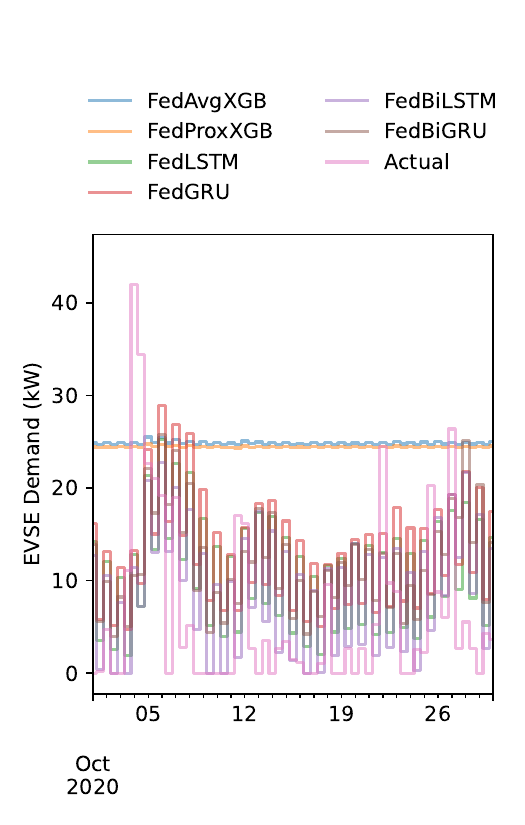}}
            \caption{Energy Demand Forecasting on a randomly selected EVSE from the test set of the Dundee (a), FEUP (b), Boulder (c), and Palo Alto (d) Datasets using FL-based methods.}
            \label{fig:actual-vs-fededf-results}
        \end{figure*}
        
        Similar conclusions can be drawn from the experimental results on the FEUP dataset, where XGBoost (cf. Figure \ref{subfig:ResultsMLFEUP}) maintains dominance over BiLSTM and statistical methods (cf. Figure \ref{subfig:ResultsBaselinesFEUP}) across most metrics. However, the elevated MASE and SMAPE scores indicate a lower prediction fidelity compared to the Dundee dataset. This can (partially) be attributed to the increased sparsity in EVSE demand, inhibiting models' ability to capture consistent temporal patterns. This hypothesis is supported by the increased MAE and decreased $R^2$ values.

        In the Boulder dataset, XGBoost outperforms all models, achieving the lowest MASE score, with a MAE range of 1.90 -- 4.16 kW, reflecting minimal deviations in demand estimation. Its SMAPE and moderate $R^2$ score suggest balanced performance in symmetric errors and trend alignment. GRU serves as the runner-up, surpassing statistical models by up to $35\%$ in terms of MASE and marginally reducing RMSE in the 75th quartile. However, its higher SMAPE indicates challenges in adapting to abrupt demand shifts, while the slightly improved $R^2$ reflects stable but limited forecasting fidelity, likely due to the intermittent patterns of the dataset.

        XGBoost remains dominant in the Palo Alto dataset as well, having the lowest MASE, with a MAE range of 5.61 -- 8.68 kW, alongside strong SMAPE and improved $R^2$ scores. Here, BiGRU follows as the runner-up, outperforming statistical models up to $62.63\%$ in terms of MASE, and reducing RMSE and MAE up to $6.39\%$ in the 25th quartile. However, its higher SMAPE and broader $R^2$ range highlight its struggles with sudden demand shifts and variable forecasting accuracy, attributed to the lumpy demand patterns of the dataset.

    \subsection{Experimental Results, Part II: Federated Learning}\label{subsec:FederatedModels}  
        \begin{table*}[!ht]
    \caption{Prediction error ($25^{th}; 50^{th}; 75^{th}$ quartile) for each FL-enabled methodology per dataset (``heavy'' configuration; MASE--MAE - lower is better; $R^2$ - higher is better).}
    \label{tab:fededf-results-heavy}
    \centering    
    \renewcommand{\arraystretch}{1.5}
    \resizebox{\textwidth}{!}{%
        \begin{tabular}{@{}lllllllll@{}}
            \toprule
            & \multirow{3}{*}{Model} 
            & \multicolumn{7}{c}{Metric}
            \\ \cmidrule(l){3-9} 
            & ~
            & MASE (\%) 
            & SMAPE (\%) 
            & MAAPE (rads) 
            & WAPE (\%) 
            & RMSE (kW) 
            & MAE (kW) 
            & $R^2$
            \\ \midrule
            \multirow{6}{*}{\rotatebox[origin=c]{90}{Dundee}}
            & FedAvgXGB
            & 0.17; 0.32; 0.43 
            & 1.91; 1.95; 1.97 
            & 1.49; 1.52; 1.53 
            & 1.07; 1.65; 2.50
            & 2.66; 4.08; 6.37 
            & 0.64; 1.04; 1.75 
            & -4.43; -0.95; 0.21 
            \\
            & FedProxXGB       
            & 0.18; 0.32; 0.43 
            & 1.91; 1.95; 1.97 
            & 1.49; 1.52; 1.53 
            & 1.09; 1.67; 2.52 
            & 2.64; 4.11; 6.41 
            & 0.64; 1.05; 1.76 
            & -4.44; -1.01; 0.2 
            \\
            & FedLSTM      
            & 0.23; 0.33; 0.46 
            & 1.88; 1.91; 1.94 
            & 1.47; 1.48; 1.51 
            & 1.23; 1.72; 2.55 
            & 1.37; 2.88; 5.25 
            & 0.58; 1.21; 2.39 
            & -0.04; 0.13; \textbf{0.33} 
            \\
            & FedGRU
            & 0.19; 0.29; 0.39 
            & 1.88; 1.90; 1.95 
            & 1.46; 1.48; 1.51 
            & 1.16; 1.52; 2.12 
            & 1.41; 2.92; 5.28 
            & 0.51; 1.13; 1.95 
            & -0.11; 0.13; 0.33 
            \\
            & FedBiLSTM       
            & 0.21; 0.31; 0.39 
            & \textbf{1.78; 1.87; 1.93} 
            & \textbf{1.38; 1.45; 1.51} 
            & \textbf{1.07}; 1.62; 2.32 
            & 1.06; \textbf{2.75}; 5.29 
            & 0.60; 0.99; 1.84 
            & -0.05; 0.16; 0.32
            \\
            & FedBiGRU          
            & \textbf{0.17; 0.27; 0.37} 
            & 1.89; 1.93; 1.96 
            & 1.47; 1.50; 1.52 
            & 1.15; \textbf{1.43; 1.88} 
            & \textbf{1.06}; 2.77; \textbf{5.15} 
            & \textbf{0.39; 0.98; 1.70}
            & \textbf{0.00; 0.16}; 0.31 
            \\ \bottomrule
            \multirow{6}{*}{\rotatebox[origin=c]{90}{FEUP}}
            & FedAvgXGB
            & \textbf{0.62}; 0.74; 0.90
            & 1.42; 1.49; 1.63 
            & 1.08; 1.12; 1.23 
            & \textbf{0.79; 0.89}; 1.12 
            & 2.75; 4.31; \textbf{6.67} 
            & \textbf{1.72}; 2.70; \textbf{4.18} 
            & \textbf{-0.01}; 0.02; 0.15 
            \\
            & FedProxXGB
            & 0.63; \textbf{0.73; 0.90}
            & 1.43; 1.51; 1.62 
            & 1.08; 1.11; 1.23 
            & 0.80; 0.90; 1.09 
            & \textbf{2.75; 4.22}; 6.80
            & 1.74; \textbf{2.61}; 4.21 
            & -0.02; \textbf{0.04; 0.20}
            \\
            & FedLSTM     
            & 0.93; 1.04; 1.17 
            & 1.40; 1.43; 1.55 
            & 1.08; 1.10; 1.16 
            & 1.00; 1.08; 1.22 
            & 3.03; 5.27; 6.98 
            & 2.63; 3.38; 4.69 
            & -0.02; 0.02; 0.08 
            \\
            & FedGRU    
            & 0.69; 0.85; 0.96 
            & 1.17; 1.30; 1.35 
            & 0.86; 0.91; 0.96 
            & 0.87; 0.91; 1.00 
            & 3.05; 5.55; 7.31 
            & 2.13; 2.71; 4.29 
            & -0.08; -0.00; 0.09 
            \\
            & FedBiLSTM          
            & 0.69; 0.84; 0.94 
            & \textbf{1.10; 1.17; 1.23} 
            & \textbf{0.77; 0.80; 0.83} 
            & 0.88; 0.90; \textbf{0.97} 
            & 3.02; 5.55; 7.41 
            & 2.09; 2.65; 4.24 
            & -0.12; -0.00; 0.08 
            \\
            & FedBiGRU           
            & 0.82; 0.89; 1.04 
            & 1.45; 1.48; 1.59 
            & 1.08; 1.10; 1.17 
            & 0.92; 0.99; 1.08 
            & 2.92; 5.38; 7.20
            & 2.36; 2.99; 4.51 
            & -0.05; 0.02; 0.13
            \\ \bottomrule
            \multirow{6}{*}{\rotatebox[origin=c]{90}{Boulder}}
            & FedAvgXGB
            & 0.74; 0.99; 1.14 
            & 1.50; 1.72; 1.85 
            & 1.20; 1.34; 1.45 
            & 1.71; 2.29; 4.50
            & 6.22; 7.24; 7.87 
            & 5.81; 6.04; 6.27 
            & -0.86; -0.34; -0.10
            \\
            & FedProxXGB
            & 0.72; 0.97; 1.11 
            & 1.49; 1.72; 1.85 
            & 1.19; 1.34; 1.46 
            & 1.66; 2.24; 4.39 
            & 6.09; 7.14; 7.82 
            & 5.65; 5.91; 6.15 
            & -0.79; -0.30; -0.08 
            \\
            & FedLSTM     
            & 0.63; 0.74; 0.94 
            & \textbf{1.27}; 1.64; 1.80
            & \textbf{1.02}; 1.28; 1.41 
            & 1.54; 2.06; 3.40
            & 5.61; 6.68; \textbf{7.30} 
            & 4.26; 5.05; 6.09 
            & -0.90; -0.30; -0.07 
            \\
            & FedGRU    
            & 0.60; 0.71; 0.93 
            & 1.41; 1.60; 1.66 
            & 1.11; 1.23; \textbf{1.28} 
            & 1.64; 2.14; 3.08 
            & \textbf{5.35}; 6.91; 7.77 
            & 3.82; 4.55; 6.15 
            & -0.70; -0.33; -0.09 
            \\
            & FedBiLSTM          
            & 0.73; 0.82; 1.01 
            & 1.30; 1.63; 1.82 
            & 1.07; 1.29; 1.43 
            & 1.90; 2.30; 3.81 
            & 6.15; 7.05; 7.93 
            & 5.30; 5.52; 6.40
            & -1.42; -0.43; -0.21 
            \\
            & FedBiGRU           
            & \textbf{0.58; 0.69; 0.85} 
            & 1.43; \textbf{1.56; 1.65} 
            & 1.14; \textbf{1.22}; 1.29 
            & \textbf{1.46; 1.86; 2.90}
            & 5.40; \textbf{6.59}; 7.36 
            & \textbf{3.48; 4.44; 5.56} 
            & \textbf{-0.55; -0.21; -0.05} 
            \\ \bottomrule
            \multirow{6}{*}{\rotatebox[origin=c]{90}{Palo Alto}}
            & FedAvgXGB
            & 1.28; 1.61; 1.90
            & 1.18; 1.41; 1.56 
            & 1.06; 1.21; 1.30
            & 1.73; 2.53; 3.91 
            & 19.34; 20.69; 21.72 
            & 17.45; 19.25; 20.72 
            & -5.76; -2.39; -1.42 
            \\
            & FedProxXGB
            & 1.26; 1.57; 1.87 
            & 1.18; 1.41; 1.56 
            & 1.06; 1.21; 1.30
            & 1.70; 2.48; 3.84 
            & 19.07; 20.31; 21.42 
            & 17.14; 18.88; 20.35 
            & -5.52; -2.29; -1.35 
            \\
            & FedLSTM     
            & 0.54; 0.68; \textbf{0.81} 
            & 1.06; 1.29; 1.40
            & 0.90; 1.06; 1.11 
            & \textbf{0.92}; 1.26; 1.54 
            & 9.03; 10.36; \textbf{11.33} 
            & 7.08; 8.08; 9.27 
            & -0.34; -0.08; \textbf{0.12} 
            \\
            & FedGRU    
            & 0.56; 0.68; 0.81 
            & 1.06; 1.27; 1.45 
            & 0.90; 1.04; 1.18 
            & 0.93; 1.20; 1.61 
            & \textbf{8.78}; 10.35; 12.08 
            & 6.90; 7.96; 9.95 
            & -0.44; \textbf{-0.07}; 0.10
            \\
            & FedBiLSTM          
            & \textbf{0.49; 0.63}; 0.84 
            & \textbf{1.05; 1.20; 1.25}
            & \textbf{0.88; 0.95; 1.01}
            & 0.93; \textbf{1.15; 1.45}
            & 9.06; 10.19; 11.54 
            & \textbf{6.71; 7.77; 8.96}
            & \textbf{-0.25}; -0.13; 0.11 
            \\
            & FedBiGRU           
            & 0.53; 0.70; 0.88 
            & 1.06; 1.27; 1.42 
            & 0.89; 1.03; 1.18 
            & 0.94; 1.21; 1.68 
            & 9.18; \textbf{10.08}; 12.05 
            & 6.76; 8.16; 9.92 
            & -0.33; -0.18; 0.09
            \\ \bottomrule
        \end{tabular}
    }
\end{table*}

        In this section, we describe the FL implementation for each model and present the results of our experimental comparison. As outlined in Section \ref{subsec:FLCompliance}, XGBoost, (Bi)LSTM, and (Bi)GRU models are compatible with FL. 

        For the FL implementation of XGBoost, we adopt FedXGBllr \cite{DBLP:conf/eurosys/MaQBL23}, a novel horizontal FL framework that treats the learning rates of the aggregated tree ensembles as learnable parameters. This is implemented through a small one-layer 1D Convolutional Neural Network (1D-CNN) -based model, whose inputs correspond to the prediction outcomes of all trees from the aggregated ensemble. The kernel and stride size of the 1D-CNN are equal to the number of trees, $M$, per client, allowing each channel to represent learnable learning rates.

        The process involves clients initially training local XGBoost tree ensembles, which are then sorted, aggregated and broadcast by the server, followed by clients collaboratively training the 1D-CNN model with the prediction outcomes of all trees from these aggregated ensembles on their local data samples serving as inputs. 
        To keep communication overhead manageable and to facilitate comparison with the centralized implementation described in Section \ref{subsec:CentralizedModels}, we set $M=37$ estimators per client. Consequently, the aggregated model ensemble will contain 296 trees for the Dundee, Boulder, and Palo Alto federations ($N = 8$ clients) and 148 trees for the FEUP federation  ($N = 4$ clients), respectively. Due to differences between FL models' architecture and methodology, we use $R' = 40$ communication rounds with $E' = 10$ local epochs and a patience of $3$ epochs.

        Figure \ref{fig:actual-vs-fededf-results} illustrates the performance of the global (aggregated) FedEDF models over the same EVSEs used in the previous sections of our experimental study, whereas the completed results are listed in the respective rows in Table \ref{tab:fededf-results-heavy}. While \cite{DBLP:conf/eurosys/MaQBL23} employs FedAvg for training the 1D-CNN model, our experiments also evaluate a FedProx variant to assess its impact, in terms of prediction fidelity.
         
% \usepackage{booktabs}
\begin{table*}[!ht]
    \caption{
        Energy consumption and carbon-equivalent emissions of centralized vs. federated EDF model variants (lower is better). The ``heavy'' FedEDF variant performs five local epochs, whereas the ``light'' FedEDF variant performs one local epoch per FL round, respectively.
    }    
    \label{tab:fededf_energy_consumption}
    \centering    
    \renewcommand{\arraystretch}{1.5}
    \resizebox{\textwidth}{!}{%
        \begin{tabular}{@{}lllcccccccc@{}}
            \toprule
            & \multirow{4.2}{*}{Model} 
            & \multicolumn{3}{c}{Total energy consumption}
            & \multicolumn{6}{c}{Federated vs. centralized EDF}
            \\ \cmidrule(l){3-11} 
            &
            & \multirow{2.5}{*}{\renewcommand{\arraystretch}{1.1}\begin{tabular}{@{}c@{}}Centralized ($E_{\text{cent}}$) \\ (kJ / epoch)\end{tabular}}
            & \multicolumn{2}{c}{\renewcommand{\arraystretch}{1.1}\begin{tabular}{@{}c@{}}Federated ($E_{\text{fed}}$) \\ (kJ / client / FL round)\end{tabular}}
            & \multicolumn{2}{c}{\renewcommand{\arraystretch}{1.1}\begin{tabular}{@{}c@{}}Log-scale energy overhead \\ $\left( log_{10}{\left( E_{\text{fed}} / E_{\text{cent}} \right)} \right)$\end{tabular}}
            & \multicolumn{2}{c}{\renewcommand{\arraystretch}{1.1}\begin{tabular}{@{}c@{}}Annual overheads \\ ($\times 10^{-3}$ kWh / FL round)\end{tabular}}
            & \multicolumn{2}{c}{\renewcommand{\arraystretch}{1.1}\begin{tabular}{@{}c@{}}CO$_2$-equivalent emissions \\ (g CO$_2$e)\end{tabular}}
            \\ \cmidrule(l){4-11} 
            & ~
            & ~
            & heavy
            & light
            & heavy
            & light
            & heavy
            & light
            & heavy
            & light
            \\ \midrule
            \multirow{5}{*}{\rotatebox[origin=c]{90}{Dundee}}
            & XGBoost      
            & \textbf{0.01}
            & \textbf{0.55}
            & \textbf{0.18}
            & 2.65
            & 2.16
            & \textbf{1.22}
            & 0.39
            & \textbf{0.35}	
            & 0.11
            \\
            & LSTM
            & 0.48
            & 1.94
            & 0.18
            & 1.51
            & \textbf{0.48}
            & 4.18
            & \textbf{0.27}
            & 1.21
            & \textbf{0.08}
            \\
            & GRU
            & 0.49
            & 1.29
            & 0.28
            & \textbf{1.32}
            & 0.66
            & 2.73
            & 0.49
            & 0.79
            & 0.14
            \\
            & BiLSTM
            & 0.72
            & 2.87
            & 0.90
            & 1.51
            & 1.00
            & 6.19
            & 1.81
            & 1.79
            & 0.52
            \\
            & BiGRU      
            & 0.65
            & 3.21
            & 0.44
            & 1.60
            & 0.74
            & 6.96
            & 0.81
            & 2.01
            & 0.23
            \\ \bottomrule
        \end{tabular}
    }
\end{table*}

        Although XGBoost achieves the best overall performance in the centralized setting across all datasets, the optimal FL model varies from dataset to dataset.
        
        In Dundee, FedBiGRU has the best MASE and WAPE scores, however its median MAE and SMAPE scores are consistently lower (30.67\%, and 7.22\%, respectively) compared to the centralized variant. In FEUP, FedProxXGB delivers the best MASE and $R^2$ scores (75\textsuperscript{th} percentile), nearly matching the centralized XGBoost, however with considerably higher MAE. For Boulder, FedBiGRU yields the best MASE, $R^2$ and MAE (3.48 -- 5.79 kW) scores, however its $R^2$ score remains significantly lower compared to the centralized BiGRU. In Palo Alto, FedBiLSTM achieves the best overall MASE and SMAPE scores, trailing its centralized variant by 15.07\% and 11.35\%, respectively, at the 75\textsuperscript{th} percentile. Compared to its centralized variant, the improved SMAPE and WAPE (up to 2.68\%) scores indicate higher prediction fidelity, however the wider $R^2$ range reveals inconsistent performance across clients/EVSE hubs.
                
        The comparison between centralized and federated results highlights a clear pattern: centralized XGBoost remains the dominant model across all datasets, whereas in FL the best model shifts to a neural network (BiGRU or BiLSTM) for Dundee, Boulder, and Palo Alto, and remains XGBoost for FEUP. 
        The performance gap is most pronounced for Dundee and Boulder. Here, centralized XGBoost outperforms the best federated model by up to 58.82\% in MASE and 228.57\% in $R^2$ (50th percentile). 

        For FEUP and Palo Alto the difference narrows. The centralized model still outperforms the federated models in MASE and $R^2$ (50th percentile) for both datasets. However, in the FEUP dataset the federated models marginally surpass the centralized one, achieving up to 3.54\% higher WAPE and 11.11\% higher $R^2$ (75\textsuperscript{th} percentile). In the Palo Alto dataset, the federated models trail behind by 34.26\% in WAPE and 73.17\% in $R^2$ (75\textsuperscript{th} percentile). These results demonstrate that FL can approximate centralized performance on some datasets, but the degree of approximation varies considerably across them.

        These results underscore the importance of model selection in privacy-preserving deployments. Federated XGBoost (FedAvgXGB/FedProxXGB) is attractive for its simplicity and close alignment with centralized training, yet it remains inferior to centralized XGBoost on all datasets, especially when data heterogeneity is low and communication costs are negligible. In contrast, for datasets exhibiting strong non-IID characteristics (e.g., Boulder, Palo Alto), federated neural-network models (FedBiGRU/FedBiLSTM) can narrow the performance gap by exploiting local sequence dynamics that are difficult to capture with a tree-based ensemble under federated constraints. Thus, a practical strategy would be to use federated XGBoost when the data share many common features and the participants are alike, and switch to federated bi-directional RNNs when the data are highly heterogeneous. Nevertheless, if privacy is not a concern, the centralized XGBoost still delivers the highest accuracy.

    \subsection{A Note on the Energy Consumption and Carbon Emission Savings}\label{subsec:FL-Energy-Footprint}
        In centralized learning frameworks, large amounts of raw data must be transferred from edge devices to a central server for training, resulting in additional bandwidth and energy for data transmission and storage. In contrast, FL frameworks train in-situ, leveraging edge devices' processing power and sharing model updates rather than raw data. By reducing reliance on large-scale centralized infrastructure(s) and maximizing the use of local resources, FL promotes energy-efficient ML while improving privacy and scalability. 
        
        By using off-the-shelf energy profiling tools, we are able to estimate the end-to-end energy overhead of FedEDF and identify opportunities for improving energy consumption without compromising model performance. For the purposes of this study, we employed the GREEN.DAT.AI Benchmark tool by INESC-TEC (patent pending)\footnote{\url{https://enerframe-greendatai.haslab-dataspace.pt/docs/index.html}}. This is the de-facto reference framework within the GREEN.DAT.AI project and has already been validated on the same FL stack, providing high-resolution per-node, per-epoch power traces while incurring virtually no measurement overhead. By analysing the profiling results, we discovered that a significant share of the power draw stemmed from the number of local epochs executed in each FL round. Consequently, we revised the FL workflow of FedEDF by adjusting this parameter. This change lowers the overall computational load, cuts energy consumption, while preserving forecasting fidelity as much as possible.

        To evaluate the impact of this optimization, we compare the total energy consumption of the centralized EDF (c.f., Section \ref{subsec:Methodology}) models against their corresponding FL-enabled variants (c.f., Section \ref{subsec:MethodologyFL}) in two configurations, namely  ``heavy'' and ``light''. The difference between the latter two configurations is the number of local epochs within an FL round, which in the ``heavy'' is set to five epochs, whereas in the ``light'' is set to one epoch. To ensure consistency across experiments, we allocate the same hardware in each run, specifically, two CPUs per client and per server instance.
    
        \begin{table*}[!ht]
    \caption{Prediction error ($25^{th}; 50^{th}; 75^{th}$ quartile) for each FL-enabled methodology on the Dundee dataset (``light'' configuration; MASE--MAE - lower is better; $R^2$ - higher is better).}
    \label{tab:fededf-results-light}
    \centering    
    \renewcommand{\arraystretch}{1.5}
    \resizebox{\textwidth}{!}{%
        \begin{tabular}{@{}lllllllll@{}}
            \toprule
            & \multirow{3}{*}{Model} 
            & \multicolumn{7}{c}{Metric}
            \\ \cmidrule(l){3-9} 
            & ~
            & MASE (\%)
            & SMAPE (\%) 
            & MAAPE (rads) 
            & WAPE (\%) 
            & RMSE (kW) 
            & MAE (kW) 
            & $R^2$
            \\ \midrule
            \multirow{6}{*}{\rotatebox[origin=c]{90}{Dundee}}
            & FedAvgXGB      
            & 0.31; 0.48; 0.75 
            & \textbf{1.13; 1.31; 1.59} 
            & \textbf{0.89; 1.02; 1.26}
            & \textbf{1.22}; 3.27; 6.12 
            & 2.75; 4.12; 6.20
            & 1.16; 1.96; 3.22 
            & -6.01; -1.11; 0.28 
            \\
            & FedProxXGB      
            & 0.32; 0.49; 0.76 
            & 1.13; 1.31; 1.59 
            & 0.89; 1.02; 1.26 
            & 1.23; 3.29; 6.15 
            & 2.76; 4.14; 6.22 
            & 1.18; 1.98; 3.26 
            & -5.99; -1.13; 0.28 
            \\
            & FedLSTM
            & 0.27; 0.36; 0.54 
            & 1.70; 1.81; 1.91 
            & 1.32; 1.40; 1.49 
            & 1.30; 2.00; 2.99 
            & 1.39; \textbf{2.73}; 5.33 
            & 0.63; 1.30; \textbf{2.62} 
            & -0.07; 0.08; 0.27 
            \\
            & FedGRU
            & 0.31; 0.42; 0.56 
            & 1.87; 1.91; 1.95 
            & 1.45; 1.48; 1.52 
            & 1.44; 2.31; 3.53 
            & 1.51; 2.91; \textbf{5.27} 
            & 0.76; 1.49; 3.13 
            & -0.11; 0.04; 0.28 
            \\
            & FedBiLSTM
            & 0.39; 0.50; 0.65 
            & 1.88; 1.91; 1.94 
            & 1.47; 1.49; 1.52 
            & 1.82; 2.68; 3.96 
            & 1.54; 2.86; 5.54 
            & 0.90; 1.65; 3.07 
            & -0.29; 0.02; 0.21
            \\
            & FedBiGRU      
            & \textbf{0.20; 0.32; 0.45} 
            & 1.49; 1.78; 1.89
            & 1.15; 1.38; 1.48 
            & 1.23; \textbf{1.61; 2.21}
            & \textbf{1.36}; 2.79; 5.35 
            & \textbf{0.46; 1.30}; 2.67
            & \textbf{0.00; 0.14; 0.29} 
            \\ \bottomrule
        \end{tabular}
    }
\end{table*}

        Table \ref{tab:fededf_energy_consumption} shows the total energy consumption of FedEDF during training, averaged over ten independent runs on the Dundee dataset, with the other three EVSE datasets sharing similar relative trends. Energy consumed by the communication network (e.g., Wi-Fi/Ethernet transmission of model updates) was not measured and therefore is not included in the numbers. It can be observed that in the centralized setting, the ``greener'' algorithm is XGBoost, which needs only $0.01$ kJ per epoch, $\approx 98$\% less than the RNNs. Training the same models under the Federated (``heavy'') configuration, XGBoost still reports the lowest energy consumption (0.55 kJ per client per round). The other RNNs (GRU, BiLSTM and BiGRU) consume considerably (up to $6 \times$) more energy. Similar observations are found for the federated (``light'') configuration, though the gap between the RNNs and XGBoost is slightly smaller, up to $5 \times$. 

        To further evaluate the energy cost of each model variant, we compute the base-10 logarithm of the federated-to-centralized EDF ratio. This metric provides a continuous measure of order-of-magnitude differences while remaining independent of absolute energy values. The pronounced log-scaled energy overhead of FL-enabled XGBoost, up to $2.7$ units higher than its centralized counterpart (vs. $1.6$ for the RNNs), stems from the FedXGBllr algorithm \cite{DBLP:conf/eurosys/MaQBL23}, where in addition to local XGBoost training, the clients collaboratively train an 1D-CNN model for collective decision making across participants. On the other hand, in the Federated (``light'') setting, the energy consumption drops dramatically (up to $68.21$\%) and becomes comparable to centralized learning, especially in the case of the LSTM where the log-scaled energy overhead of its FL-enabled variant is just $0.48$ units higher than its centralized variant. Regardless, all RNNs remain consistently more energy-consuming across both federated configurations.

        These observations are consistent with the findings of Qiu et al. \cite{DBLP:journals/jmlr/QiuPFGGBTML23}, who report that, depending on the configuration, federated learning can emit up to two orders of magnitude more carbon than centralized training, while in certain cases it can be comparable to centralized learning because of the reduced energy consumption of embedded devices and/or careful tuning of the FL configuration.

        The energy overheads of FedEDF can also be expressed in terms of annual CO\(_2\)-equivalent emisions that result from the energy demand during model training \cite{PSOMOPOULOS2010485}. Focusing (without loss of generalization) on the FL-enabled XGBoost model and assuming that the model is re-trained periodically (e.g., once per year), the average required energy (in kWh) per run is equals to $E_{\text{run, kWh}}^{\text{heavy}} = 1.23 \times 10^{-3}$ kWh, and $E_{\text{run, kWh}}^{\text{light}} = 0.39 \times 10^{-3}$ kWh, for the ``heavy'', and ``light'' configuration, respectively.

        Since the EVSEs of the Dundee dataset are located in Scotland, UK, we use the average value of the ``Greenhouse gas emission intensity of electricity generation in Europe'' index\footnote{Greenhouse gas emission intensity of electricity generation in Europe, \url{https://www.eea.europa.eu/en/analysis/indicators/greenhouse-gas-emission-intensity-of-1}. Last visited: 15 December 2025} for the closest available year, or:

        \begin{equation}
            EF_{\text[grid, 2018]}^{\text{EU-27}} = 0.289\ kg\ CO_2 e / kWh
        \end{equation}

        \noindent
        Applying this factor over the energy difference with respect to its corresponding centralized (baseline) configuration, the annual greenhouse gas emissions overhead equals to:

        \begin{equation}
            CO_{\text{2, 2018}}^{\text{baseline, algo}} = \Delta E_{\text{run}}^{\text{baseline, algo}} \times EF_{\text[grid, 2018]}^{\text{EU-27}}
        \end{equation}

        \noindent 
        or, $CO_{\text{2, 2018}}^{\text{baseline, heavy}} = 0.35$ $g\ CO_2 e$, and $CO_{\text{2, 2018}}^{\text{baseline, light}} = 0.11$ $g\ CO_2 e$ for the ``light'' and ``heavy'' configurations of FedEDF, respectively (see the ``CO$_2$ equivalent'' column of Table \ref{tab:fededf_energy_consumption}). In general, it can be observed that the ``light'' configuration reduces additional greenhouse-gas emissions by $80.6$\%, in average, thus reinforcing the impact of the ``light'' configuration for positioning FedEDF as an environmentally responsible solution for smart urban infrastructure and planning.

        To assess the energy consumption vs. performance trade-off, Table \ref{tab:fededf-results-light} illustrates the performance of each FL-enabled model in the ``light'' configuration for the Dundee dataset. Compared to the ``heavy'' configuration, it can be observed that FedBiGRU remains the better model, albeit with $12.50$\% and $0.72$\% loss in prediction fidelity, in terms of median $R^2$, and $RMSE$ scores, respectively. The SMAPE score, however, remains approximately consistent, therefore the prediction quality is reduced only in terms of energy demand, as further supported by the increased MAE score ($0.46 - 2.67$ kW; up to $57.06$\%). Similar reductions in prediction fidelity are observed in the other RNNs, and the XGBoost as well. This behavior is attributed to the number of local epochs which reduced energy consumption, but introduced larger deviation between the participants' local model, which in turn impacts the generalization of the (global)FedEDF model. 
        
        As observed in the previous comparison, increasing the number of local epochs and/or FL rounds can alleviate this behavior, and converge on a better global model. However, this comes at a cost of deceased energy efficiency. Therefore, a potential area of improvement is to focus on minimizing (local)FedEDF deviation, either by adjusting the $mu_{prox}$ parameter and/or combining it with other algorimths such as FedNova \cite{DBLP:conf/nips/WangLLJP20}, or SCAFFOLD \cite{DBLP:conf/icml/KarimireddyKMRS20}.

%% Conclusion
\section{Conclusion}\label{sec:conclusion}      
    In this paper, we present an experimental comparison of well-established time series forecasting methods for the EDF problem. We evaluate the pros and cons of each approach and, in addition, we assess the impact of Federated Learning on this task. Through an extensive study using four real-world EVSE data, we analyze the performance of these methods and highlight the trade-offs between forecasting accuracy, privacy, and energy consumption associated with Federated Learning.

    In the near future, we aim to further extend our efforts on feature engineering by including additional factors, related to driving behavior and/or spatial characteristics of the region.  
    In a parallel line of research, and as our long-term goal, we aim to experiment with other FL aggregation strategies, in order to exploit on the EVSE topology, rather than treating each charging station as independent of the others.

\section*{Acknowledgment}
    \noindent
    This work was supported in part by the Horizon Europe Research and Innovation Programme of the European Union under grant agreement No. 101070416 (Green.Dat.AI; \url{https://greendatai.eu}). In this work, INESC TEC provided the requirements of the business case, as well as access to the FEUP dataset and their Energy benchmarking tool.

\bibliographystyle{IEEEtran.bst}
\bibliography{bibliography.bib}

\end{document}